\documentclass[11pt,letterpaper]{amsart}
\usepackage{amsaddr}

\usepackage{amssymb,latexsym,amsfonts,amsmath}
\usepackage{geometry}
\usepackage[authoryear,sort&compress]{natbib}
\bibpunct{[}{]}{;}{n}{,}{,}

\usepackage{hyperref}
\usepackage{booktabs}
\usepackage{comment}
\usepackage[shortlabels]{enumitem}
\usepackage{mathrsfs}
\usepackage{amsthm}
\usepackage{tikz}
    \usetikzlibrary{matrix}
    \usetikzlibrary{decorations.markings}
    \usetikzlibrary{arrows}
\usepackage{hyperref}
\usepackage[T1]{fontenc}
\usepackage{oldgerm}
\usepackage{scalerel,stackengine}
\usepackage{stmaryrd}
\usepackage{xcolor}
\usepackage{xfrac}
\usepackage{multirow}
\usepackage{makecell}
\usepackage{subfig}
\usepackage{svg}
\svgsetup{inkscapelatex=false}
\usepackage[capitalise]{cleveref}
\newtheorem{theorem}{Theorem}[section]
\newtheorem{definition}{Definition}[section]

\newtheorem{lemma}{Lemma}[section]

\newtheorem{remark}{Remark}[section]
\newtheorem{hypothesis}{Hypothesis}[section]

\newcommand{\x}{{\rm x}}

\newcommand{\bx}{\mathbf{x}}
\newcommand{\y}{{\rm y}}
\newcommand{\by}{\mathbf{y}}
\newcommand{\z}{{\rm z}}
\newcommand{\bz}{\mathbf{z}}

\begin{document}

\title[]{The asymptotic behavior of attention in transformers}

\author{\'A. Rodr\'iguez Abella$^1$, J.P. Silvestre$^2$, P. Tabuada$^2$}
\address{$^1$Department of Applied Mathematics, ICAI School of Engineering, Comillas Pontifical University, Madrid, Spain}\address{$^2$Department of Electrical and Computer Engineering, University of California, Los Angeles, USA }\email{arabella@comillas.edu, joaosilvestre@g.ucla.edu, tabuada@ee.ucla.edu}

\begin{abstract}
The transformer architecture has become the foundation of modern Large Language Models (LLMs), yet its theoretical properties are still not well understood. As with classic neural networks, a common approach to improve these models is to increase their size and depth. However, such strategies may be suboptimal, as several works have shown that adding more layers yields increasingly diminishing returns. More importantly, prior studies have shown that increasing depth may lead to model collapse, \textit{i.e.}, all the tokens converge to a single cluster, undermining the ability of LLMs to generate diverse outputs. Building on differential equation models for the transformer dynamics, we prove that all the tokens in a transformer asymptotically converge to a cluster as depth increases. At the technical level we leverage tools from control theory, including consensus dynamics on manifolds and input-to-state stability (ISS). We then extend our analysis to auto-regressive models, exploiting their structure to further generalize the theoretical guarantees. 
\end{abstract}

\maketitle

\section{Introduction}

The transformer architecture~\cite{vaswani2017attention}, introduced in 2017, was a groundbreaking work that established the foundation for current Large Language Models (LLMs). In contrast to older recurrent architectures~\cite{hochreiter1997long,rumelhart1985learning}, transformers were primarily based on the attention mechanism~\cite{chorowski2015attention}, making them significantly more efficient and powerful. Since then, many works have strived to improve the architecture through modifications, such as rotary embeddings~\cite{su2024roformer}, although the core structure has remained intact over the years.

Nearly a decade later, this architecture is still not fully understood~\cite{van2023large}. Numerous works have attempted to bridge the gap between experimental results and theoretical guarantees, some focusing on the approximation capabilities of transformers~\cite{yun2019transformers}, others on their in-context behavior~\cite{garg2022can}. Beyond simply analyzing or corroborating empirical results, many theoretical efforts have aimed to justify or guide the evolution of transformers by studying individual components such as the attention block~\cite{dong2021attention}.

Although several theoretical results suggest the key to better transformers may not just be depth~\cite{levine2020limits}, the trend toward ever-larger models continues. In fact, recent commercial models, such as the LLaMA 3 405B, have surpassed one hundred layers~\cite{grattafiori2024llama}. This increase in depth, along with the significant number of parameters, has made recent LLMs prohibitively expensive to train and operate~\cite{cottier2024rising}.  As the cost of transformers continues to rise, this trend underscores a pressing question: does increasing the number of layers actually improve performance?

As more works attempt to answer this question, one conclusion becomes increasingly clear: deeper is not always better. In fact, some empirical studies on recent models suggest that the final layers do not significantly contribute to the model’s representational capacity~\cite{csordas2025language,petty2023impact}. These studies corroborate theoretical results showing that the expressive power of transformers increases with depth only up to a certain threshold~\cite{levine2020limits}. Furthermore, as highlighted by~\cite{wang2024deepnet}, training transformers beyond a few hundred layers is often difficult unless architectural modifications are added.

Other works reached a more dooming conclusion: as depth increases, tokens begin to cluster. One of the first studies on the topic~\cite{dong2021attention}, proved that pure attention loses rank with depth. In such event the model may collapse, i.e., lose the ability to generate diverse outputs, as the depth increases. However, a common limitation of these works typically lies in the strong assumptions that limit their practical implications. Thus, some studies have used more general models, considering for instance the norm in each layer \cite{wu2024role}.

Our work focuses on expanding these results by relaxing some of their assumptions, leveraging well-established results from control theory\textit{, e.g.,} consensus~\cite{sarlette2009consensus,kraisler2023consensus} and input-to-state stability (ISS)~\cite{Sontag2008}. We build on the mathematical model derived in~\cite{geshkovski2023mathematical}, which describes the transformer dynamics by a differential equation, and prove that all tokens converge to a cluster. By interpreting tokens as particles on the sphere, we leverage the extensive literature on consensus dynamics defined over such manifolds~\cite{markdahl2017almost}. Additionally, using ISS theory, we extend the results of~\cite{geshkovski2023mathematical,geshkovski2024dynamic} to auto-regressive models such as GPT-2, while allowing the weight matrices to vary with depth, unlike previous works. Finally, our theoretical predictions are confirmed through experiments on two popular models: GPT-2 and GPT-Neo.
\begin{table*}[ht]
\caption{Summary of the results presented in this work for several particular cases of the continuous model \cref{Transformer}, where $Q(t)$, $K(t)$, and $U(t)$ denote the query, key and value matrices, respectively.}
\label{table:summary}
\vskip 0.1in
\begin{center}
\begin{scriptsize}

\textbf{Full attention}\\[2pt]
\begin{tabular}{|c|c|c|}
\hline
\textbf{Section} & \cref{sec:grad} & \cref{sec:full}\\
\hline
\textbf{\# of heads} & $h=1$ & $h\geq 1$\\
\hline
$P(t)=Q(t)^\top K(t)$ & \makecell{Time invariant,\\ symmetric, positive definite} & \makecell{Time varying,\\unif. continuous, bounded}\\
\hline
$U(t)$ & Identity & Identity\\
\hline
\textbf{Result} & \cref{theorem:grad} & \cref{theorem:hemisphere}\\
\hline
\textbf{Statement} & \makecell{Gradient flow,\\ convergence to equilibrium} & \makecell{Convergence\\ to consensus}\\
\hline
\textbf{Domain of attraction} & Whole sphere & Some hemisphere\\
\hline
\end{tabular}

\vskip 0.15in

\textbf{Causal attention (auto-regressive)}\\[2pt]
\begin{tabular}{|c|c|c|c|}
\hline
\textbf{Section} & \cref{sec:masked_id} & \cref{sec:masked_symmetric} & \cref{sec:timevaryingU}\\
\hline
\textbf{\# of heads} & $h\geq 1$ & $h=1$ & $h=1$\\
\hline
$P(t)=Q(t)^\top K(t)$ & \makecell{Time varying,\\ bounded} & \makecell{Time varying,\\ bounded} & \makecell{Time varying,\\ bounded}\\
\hline
$U(t)$ & Identity & \makecell{Time invariant,\\ symmetric} & \makecell{Time varying,\\ bounded, symmetric}\\
\hline
\textbf{Result} & \cref{theorem:consensussytem} & \cref{theorem:consensussytemU} & \cref{theorem:timevaryingU}\\
\hline
\textbf{Statement} & \makecell{Asympt. stability\\ of consensus} & \makecell{Asympt. stability\\ of consensus} & \makecell{Convergence to\\ ball around consensus}\\
\hline
\textbf{Domain of attraction} & \makecell{Conull (compl.\\ of zero measure)} & Fixed hemisphere & \makecell{Time-varying\\ hemisphere}\\
\hline
\end{tabular}

\end{scriptsize}
\end{center}
\vskip -0.1in
\end{table*}

\subsection*{Contributions of the paper}

The main contribution of this paper is to provide a number of results for a differential equation model of attention\footnote{Since we focus on the attention mechanism, this model does not describe the effect of feed-forward layers.}   showing that all tokens converge to a single cluster thereby leading to a collapse of the model. We use the term \emph{consensus equilibria} to refer to such clusters as is done in the consensus literature~\cite{Survey2,Survey1}. These results hold under different assumptions on the parameters of the model ---namely, the query ($Q$), key ($K$) and value matrices ($U$), as well as the number of heads ($h$)--- that are summarized in \cref{table:summary}. 

In particular, with \cref{theorem:grad} we prove that the dynamics of the transformer is a Riemannian gradient vector field, from which we conclude convergence  to an equilibrium point (guaranteed to be of consensus type when $P=Q^\top K$ is the identity) for every initial position of the tokens. Although the gradient nature of the dynamics, in this case, was already observed and exploited in~\cite{GeLePoRi2024a}, for the benefit of the readers we provide a formal proof of this fact in a slightly more general setting. \cref{theorem:hemisphere} states that tokens converge to a consensus equilibrium whenever their starting positions lie in the interior of some hemisphere of the ellipsoid. This result holds for any number of heads and time varying  matrix $P=Q^\top K$ provided that $U$ is the identity and $P$ is bounded and uniformly continuous as a function of the time. A similar result is reported in~\cite{GeLePoRi2024a} under Lemma 4.2. However, its conclusions hold under the stronger assumptions that both $U$ and $P=Q^\top K$ are the identity matrix and there is a single attention head. 

\cref{theorem:grad,theorem:hemisphere} make no assumptions on the attention matrix other those induced by the assumptions on $P$. In contrast, \cref{theorem:consensussytem,theorem:consensussytemU,theorem:timevaryingU} focus on the auto-regressive case, also known as causal attention, where the self-attention matrix is lower triangular. \cref{theorem:consensussytem} states that when $U$ is the identity, the first token is fixed and all the other tokens converge to the position of the first one for almost every initial position of the tokens. In fact, we have asymptotic stability of this consensus equilibrium. This holds for any number of heads and any time varying $P$ matrix provided it is bounded. Similar conclusions are reported under Theorem 4.1 in~\cite{KaPoRi2024} by imposing stronger assumptions: time invariance of $P=Q^\top K$  and existence of a single attention head. \cref{theorem:consensussytemU} extends these  result to the case where $U$ is a time invariant symmetric matrix and the multiplicity of its largest eigenvalue is one. In this case all the tokens will converge to a consensus equilibrium (moreover, that equilibrium is asymptotically stable) if they start in one of the two hemispheres defined by the eigenvector associated with the largest eigenvalue of $U$. We were only able to establish this result for the single-head case although we believe it holds in greater generality. To the best of the authors' knowledge there is no result available in the literature for the case where $U$ is not the identity matrix although this is  conjectured, but not proved, in~\cite{KaPoRi2024}. Lastly, in \cref{theorem:timevaryingU} we consider the case where $U(t)$ is time varying. Thus, the eigenvector corresponding to the largest eigenvalue is also time varying, and the result states that the tokens will converge to a ball around the consensus whose radius depends on how fast this eigenvector moves. The convergence takes place provided all tokens start on the hemisphere determined by this eigenvector at the initial time. This paper extends the authors’ previous work in \cite{abellaconsensus} by incorporating two additional scenarios: the gradient flow and the time-varying value matrix. In addition, it provides complete proofs of all results.

Our theoretical findings are validated using experiments with the GPT-2 and the GPT-Neo models, providing empirical evidence for convergence to consensus equilibria in more general situations than those captured by our theoretical results. Thus, demonstrating additional confirmation for model collapse. 


\subsection*{Notations and conventions}

We use the letters $n,\ell,r,$ and $s$ to denote elements of $\mathbb{N}=\{1,2,\hdots\}$. The space of $r\times s$ real matrices is denoted by $\mathcal M_{r\times s}(\mathbb R)$. In particular, $\mathbb I_r\in\mathcal M_{r\times r}(\mathbb R)$ denotes the identity matrix. The Frobenius norm of a square matrix $A\in\mathcal M_{r\times r}(\mathbb R)$ is denoted by $\|A\|$. The elements of $\mathbb R^{n+1}$ are regarded as column matrices, i.e., $\x\in\mathbb R^{n+1}\equiv\mathcal M_{(n+1)\times 1}(\mathbb R)$. When it is convenient, tuples of $\ell$ elements of $\mathbb R^{n+1}$, $\bx=(\x_1,\hdots, \x_\ell)\in(\mathbb R^{n+1})^\ell$ (note the different font), will be regarded either as matrices, $\bx\in\mathcal M_{(n+1)\times\ell}(\mathbb R)$, or column matrices, $\bx\in\mathcal M_{(n+1)\ell\times 1}(\mathbb R)$. The tangent space of a smooth manifold $M$ at $p\in M$ and its elements are denoted by $T_pM$ and $X_p\in T_p M$, respectively. The tangent bundle and the space of vector fields of a smooth manifold $M$ are denoted by $\pi_M:TM\to M$ and $\mathfrak X(M)=\Gamma(\pi_M)$, respectively. In the same vein, the space of $k$-forms is denoted by $\Omega^k(M)$. Let $M$ be a smooth manifold. The inner product between the vectors $X_p\in T_p M$ and $Y_p\in T_p M$, according to a Riemannian metric $g$ on $M$, is denoted by $\langle X_p,Y_p\rangle_{g(p)}$, and the norm of the vector $X_p$ computed with the metric $g$ is denoted by $|X_p|_{g(p)}=\langle X_p,X_p\rangle_{g(p)}^{1/2}$. Similarly, the gradient of a function $\phi\in C^\infty(M)$ is the vector field ${\rm grad}_g\phi=( \mathbf{d}\phi)^{\sharp_g}\in\mathfrak X(M)$, where $\mathbf{d}\phi\in\Omega^1(M)$ is the exterior derivative of $\phi$ and $\sharp_g:T^*M\to TM$ denotes the sharp isomorphism, i.e., $\mathbf d\phi(X)=\langle{\rm grad}_g\phi,X\rangle_g$ for each $X\in\mathfrak{X}(M)$. In coordinates, it is given by:
\begin{equation}\label{eq:gradcoordinates}
{\rm grad}_g\phi(\bx)=g(\bx)^{-1}\frac{\partial\phi(\bx)}{\partial\bx},\qquad\bx\in M.
\end{equation}
Given another smooth manifold $N$, the tangent map of a smooth map $\phi\in C^\infty(M,N)$ is denoted by $T\phi:TM\to TN$ while its pullback is denoted by $\phi^*:T^*N\to T^*M$.

\section{Dynamics of transformers}\label{sec:model}

\subsection{Configuration space}\label{subsec:confspace}

Let $\ell,n\in\mathbb N$. A symmetric, positive-definite matrix \mbox{$W\in\mathcal M_{(n+1)\times(n+1)}(\mathbb R)$} defines an inner product on $\mathbb R^{n+1}$:
\begin{equation*}
\langle X_\x,Y_\x\rangle_W=X_\x^\top W Y_\x,
\end{equation*}
for each $X_\x,Y_\x\in T_\x\mathbb R^{n+1}$ and $\x\in\mathbb R^{n+1}$, where the superscript $\top$ denotes the transpose. The corresponding norm is denoted by $|X_\x|_W=(X_\x^\top W X_\x)^{1/2}$. The points of $\mathbb R^{n+1}$ of unit norm define an $n$-dimensional ellipsoid, which is denoted by:
\begin{equation*}
\mathcal E_W^n =\{\x\in\mathbb R^{n+1}\mid \x^\top W\x=1\}.
\end{equation*}
In this work, we consider a transformer consisting of $\ell$ tokens of dimension $n+1$ constrained to evolve on an ellipsoid. 
As we have $\ell$ tokens, the resulting state space  is the Cartesian product of $\ell$ copies of the ellipsoid, i.e.:
\begin{equation*}
(\mathcal E_W^n)^\ell=\underbrace{\mathcal E_W^n \times{\hdots}\times\mathcal E_W^n }_{\ell\text{-times}},
\end{equation*}
which is an embedded submanifold of:
\begin{equation*}
(\mathbb R_0^{n+1})^\ell=\underbrace{\mathbb R_0^{n+1}\times{\hdots}\times\mathbb R_0^{n+1}}_{\ell\text{-times}},
\end{equation*}
where $\mathbb R_0^{n+1}=\mathbb R^{n+1}-\{0\}$. The natural inclusion is denoted by $\imath_W:(\mathcal E_W^n)^\ell\hookrightarrow(\mathbb R_0^{n+1})^\ell$ and we define the projection as:
\begin{equation*}
\boldsymbol\pi_W:(\mathbb R_0^{n+1})^\ell\to(\mathcal E_W^n)^\ell,\quad\boldsymbol\pi_W=\underbrace{\pi_W\times\hdots\times\pi_W}_{\ell\text{-times}},
\end{equation*}
with $\pi_W:\mathbb R_0^{n+1}\to\mathcal E_W^n$ given by $\pi_W(\x)=\x|\x|_W^{-1}$ for each $\x\in\mathbb R_0^{n+1}$. The corresponding tangent map is readily seen to be:
\begin{equation*}
T_\bx\boldsymbol\pi_W:T_\bx(\mathbb R_0^{n+1})^\ell\to T_{\boldsymbol\pi_W(\bx)}(\mathcal E_W^n)^\ell,\quad T_\bx\boldsymbol\pi_W=T_{\x _1}\pi_W\times\hdots\times T_{\x _\ell}\pi_W,
\end{equation*}
for each $\bx=(\x_1,\hdots,\x_\ell)\in(\mathbb R_0^{n+1})^\ell$, with $T_\x\pi_W:T_\x\mathbb R_0^{n+1}\to T_{\pi_W(\x)}\mathcal E_W^n$ given by: 
\begin{equation}\label{eq:TpiW}
T_\x \pi_W\cdot X_\x =|\x|_W^{-1}\left(\mathbb I_{n+1}-\x\x^\top W|\x|_W^{-2}\right)\cdot X_{\x },
\end{equation}
for each $\x\in\mathbb R_0^{n+1}$ and $X_\x\in T_\x\mathbb R_0^{n+1}$. In particular, for $\y\in\mathcal E_W^n$ and $X_\y\in T_\y\mathcal E_W^n$, we have:
\begin{equation*}
T_\y\pi_W\cdot X_\y=\left(\mathbb I_{n+1}-\y\y^\top W\right)\cdot X_\y.
\end{equation*}

\begin{remark}[Tangent bundle of the ellipsoid]\label{remark:tangentbundles}\rm
For each $\x\in\mathbb R_0^{n+1}$, we make the identification $T_\x\mathbb R_0^{n+1}\simeq\mathbb R^{n+1}$. In particular, for $\y\in\mathcal E_W^n $, we have:
\begin{equation*}
T_\y\mathcal E_W^n =\{Y_\y\in T_\y\mathbb R_0^{n+1}\simeq\mathbb R^{n+1}\mid\y^\top W Y_\y=0\}.
\end{equation*}
Therefore, the tangent space of $(\mathcal E_W^n)^\ell$ at each $\by=(\y_1,\hdots,\y_\ell)\in(\mathcal E_W^n)^\ell$ reads:
\begin{align*}
T_\by(\mathcal E_W^n)^\ell & =T_{\y_1}\mathcal E_W^n \times{\dots}\times T_{\y_\ell}\mathcal E_W^n \\
& =\left\{Y_\by=(Y_{\y_1}, \hdots , Y_{\y_\ell})\in(\mathbb R^{n+1})^\ell\mid\y_i^\top W Y_{\y_i}=0,~1\leq i\leq \ell\right\}.
\end{align*}
\end{remark}

\begin{remark}[Evolution on the sphere]\rm
There are a number of models in which the tokens evolve on the $n$-sphere, i.e., $\mathbb S^n=\mathcal E_{\mathbb I_{n+1}}^n$. For brevity, in that case we will drop the subscripts standing for the matrix $W=\mathbb I_{n+1}$. For instance, we will write $|\cdot|=|\cdot|_{\mathbb I_{n+1}}$, $\boldsymbol\pi=\boldsymbol\pi_{\mathbb I_{n+1}}$, etc.
\end{remark}

\subsection{Discrete-time attention model} In this section we present the mathematical model for a transformer. Similarly to~\cite{GeLePoRi2024a}, the model encompasses the self-attention mechanism, the skip connection, and the normalization layer, but excludes the feedforward layer. 

Let $w\in\mathbb N$ be a design parameter. 
The weight matrices at the $k$-th layer of the transformer, $k\in\mathbb N$, are denoted by \mbox{$Q(k)\in \mathcal{M}_{w\times(n+1)}(\mathbb R)$}, \mbox{$K(k)\in \mathcal{M}_{w\times(n+1)}(\mathbb R)$} and $V(k)\in \mathcal{M}_{w\times(n+1)}(\mathbb R)$, and are typically known as the \emph{Query}, \emph{Key}, and \emph{Value}\footnote{In the introduction we used $U$ to refer to the value matrix; this difference is resolved  in this section.} matrices, respectively. 
The input to the $k$-th layer is denoted by \mbox{$\bx\in\mathcal{M}_{(n+1)\times\ell}(\mathbb{R})$} and the output \mbox{$\boldsymbol{z}\in \mathcal{M}_{w\times\ell}(\mathbb{R})$} of the self-attention mechanism is given by:
\begin{equation}
\boldsymbol{z}(k)=
V(k)\bx(k)D(k)\exp\left(\bx(k)^\top K(k)^\top Q(k)\bx(k)\right),
\label{eqmachine}
\end{equation}
where $\exp(\cdot)$ denotes the entry-wise exponential (i.e., $[\exp(R)]_{ij}=e^{R_{ij}}$), and $D(k)\in\mathcal{M}_{\ell\times\ell}(\mathbb{R})$ is defined as:
\begin{align*}
& D(k)_{ij}=
 \left(\sqrt{n+1}\sum_{l=1}^\ell\exp(\x_l(k)^\top K(k)^\top Q(k)\x_i(k))\right)^{-1},
\end{align*}
if $i=j$, and $D(k)_{ij}=0$ otherwise.

Practical transformer applications often distribute the computations of the self-attention mechanism through several parallel \emph{heads}, leading to what is commonly known as \emph{multi-headed self-attention}. 
To make explicit the dependence on the head, we write \cref{eqmachine} as:
\begin{equation*}
\boldsymbol{z}_\eta(k)=V_\eta(k)\bx(k)D_\eta(k)\exp\left(\bx(k)^\top K_\eta(k)^\top Q_\eta(k)\bx(k)\right),\label{eqmachineh}
\end{equation*}
for each $1\leq\eta\leq h$.

The outputs from all attention heads are added after being multiplied by certain weight matrices \mbox{$W_\eta\in\mathcal{M}_{(n+1)\times w}(\mathbb{R})$}, $1\leq\eta\leq\ell$. Then, the resulting sum is added to the input of the layer $\bx(k)$, using what is often called a \emph{skip connection}. Lastly, a normalization function is applied to ensure that the output is bounded. In this work, we consider functions that normalize each token of the transformer separately, which is known as \emph{layer normalization} and was first proposed in~\cite{BaKiHi2016}. 
Hence, the normalization function $\mathbf N:\mathcal{M}_{(n+1)\times\ell}(\mathbb{R})\to \mathcal{M}_{(n+1)\times\ell}(\mathbb{R})$ is of the form:
\begin{equation*}
\bx=(\x_1,\hdots,\x_\ell)\mapsto\mathbf N(\bx)=(N(\x_1),\hdots,N(\x_\ell)),
\end{equation*}
for some $N:\mathbb R^{n+1}\to\mathbb R^{n+1}$. 
Similarly to~\cite{GeLePoRi2024a}, in the following we consider the normalization function $N=\pi_W$ introduced in \cref{subsec:confspace}, which projects each token to the ellipsoid $\mathcal E_W^n $. In practice, this projection has been explicitly used in some models such as~\cite{jiang2023mistral}. For clarity, we utilize the symbol $\by=(\y_1,\hdots,\y_\ell)$ for the tokens evolving on the ellipsoid (after this explicit choice of normalization). The resulting discrete-time dynamical system reads:
\begin{align}\label{disc}
& \y_i(k+1)= \\\nonumber
& \pi_W\left(\y_i(k)+\sum_{\eta=1}^h\sum_{j=1}^\ell W_\eta (k)V_\eta(k)D_\eta(k)_{ii}\exp\left(\y_j(k)^\top K_\eta(k)^\top Q_\eta(k)\y_i(k)\right)\y_j(k)\right),
\end{align}
where $1\leq i\leq\ell$ indexes each token.

\begin{remark}[Standard layer normalization]
The standard layer normalization utilized in most transformers is given by:
\begin{equation*}
N(\x)=\frac{1}{\sigma(\x)}(\x-\mu(\x)\mathbf 1)\star\gamma+\beta,
\end{equation*}
for each $\x=(x^1,\hdots,x^{n+1})\in\mathbb R^{n+1}$. In the previous expression, $\mathbf 1$ denotes the vector $(1,\hdots,1)\in\mathbb R^{n+1}$, $\star$ denotes the element-wise product of vectors, and $\gamma,\beta\in\mathbb R^{n+1}$ are the \emph{learned scale} and \emph{shift}, respectively. Similarly, $\mu(\x)$ and $\sigma(\x)$ denote the mean and standard deviation of $\x$, respectively.
Under this normalization, we have:
\begin{align*}
|N(\x)-\beta|^2 
& =\frac{1}{\sigma(\x)^2}\sum_{\mu=1}^{n+1}(x^\mu-\mu(\x))^2(\gamma^\mu)^2\\
& =\frac{n+1}{\sum_{\mu=1}^{n+1}\left(x^\mu-\mu(\x)\right)^2}\sum_{\mu=1}^{n+1}(x^\mu-\mu(\x))^2(\gamma^\mu)^2,
\end{align*}
where we used the notation $\gamma=(\gamma^1,\hdots,\gamma^{n+1})$. It is clear that, if $\gamma=\gamma_0\mathbf 1$ for some $\gamma_0\neq 0$, then the tokens lie on the $n$-sphere of center $\beta$ and radius $(n+1)\gamma_0^2$. Therefore, the results in this paper also apply to this type of normalization. We conjecture that, for arbitrary scale parameters, the tokens will also lie on a certain hypersurface of $\mathbb R^{n+1}$ diffeomorphic to the $n$-sphere, but a more careful analysis has to be carried out to understand the geometry of such hypersurface.
\end{remark}

\subsection{Continuous-time attention model}\label{sec:continuousmodel}
In this section we introduce additional notation that is only used to derive the continuous time model. Readers not interested in the model's derivation can skip the next two paragraphs and start reading the paragraph commencing with ``To simplify notation''. 

Let $Y\in \mathfrak X((\mathcal{E}_W^n)^\ell)$ be a vector field and denote its flow by $Y^\tau:(\mathcal{E}_W^n)^\ell\to (\mathcal{E}_W^n)^\ell$. Note that $Y$ is complete, i.e., its flow is defined for each $\tau\in\mathbb R$, since $(\mathcal E_W^n)^\ell$ is compact. Given a map \mbox{$g:(\mathcal{E}_W^n)^\ell\times \mathbb{R}\to\mathbb{R}_0^+$}, we use the notation $g(\by,\tau)=O_\by(\tau^2)$ to denote the existence of a constant $T\in \mathbb{R}^+$ and a function $\sigma:(\mathcal{E}_W^n)^\ell\to \mathbb{R}_0^+$ such that, for each $\tau\in [0,T]$ and $\by\in (\mathcal{E}_W^n)^\ell$, we have $g(\by,\tau)\le \sigma(\by)\tau^2$. A map $\phi:(\mathcal{E}_W^n)^\ell\times \mathbb{R}\to (\mathcal{E}_W^n)^\ell$ is a first order approximation to the flow $Y^\tau$ if $\mathbf{d}( Y^\tau(\by),\phi(\by,\tau))= O_\by(\tau^2)$ where $\mathbf{d}$ denotes the distance on $(\mathcal{E}_W^n)^\ell$ induced by the Euclidean distance on $(\mathbb{R}^{n+1}_0)^{\ell}$.

Using the concepts introduced in the previous paragraph, our objective is to construct a vector field $Y$ such that the map defined by the right-hand side of~\cref{disc} is the best first order approximation of $Y^\tau$. To that end, we write $V_\eta(k)$ as $V_\eta(k)=\tau V_\eta'(k)$ for each $1\leq\eta\leq h$, with $0<\tau\ll1$ being a small parameter. Hence, \cref{disc} may be rewritten as:
\begin{equation*}
\y_i(k+1)=\pi_W(\y_i(k)+\tau f_k(\by(k)),\qquad 1\leq i\leq\ell,
\end{equation*}
where $f_k:(\mathcal E_W^n)^\ell\to\mathbb R^{n+1}$ is defined as:
\begin{align*}
f_k(\by)= & \sum_{\eta=1}^h\sum_{j=1}^\ell W_\eta (k)V_\eta(k)' D_\eta(k)_{ii}\exp\left(\y_j^\top K_\eta(k)^\top Q_\eta(k)\y_i\right)\y_j,
\end{align*}
for each $\by\in(\mathcal E_W^n)^\ell$. For each $1\leq i\leq\ell$, the best linear approximation in $\tau$ is given by:
\begin{align*}
\dot\y_i=\left.\frac{d}{d\tau}\right|_{\tau=0}\pi_W(\y_i+\tau f_k(\by))=T_{\y_i}\pi_W\cdot f_k(\y_i).
\end{align*}
Therefore, the continuous-time model is given by:
\begin{equation}
\label{Eq:Aux}
\dot{\y}_i
=T_{\y_i}\pi_W\cdot\left(\sum_{\eta=1}^h\sum_{j=1}^\ell W_\eta (t)V'_\eta(t)D_\eta(t)_{ii}\exp\left(\y_j^\top K_\eta(t)^\top Q_\eta(t)\y_i\right)\y_j\right),
\end{equation}
with $1\leq i\leq\ell$, $\by=(\y_1,\hdots,\y_\ell)\in(\mathcal E_W^n)^\ell$, and $t\in\mathbb R_0^+$, as the differential equation whose solution provides the best first order approximation of~\cref{disc}.

To simplify notation we introduce the following (time-dependent) auxiliary matrices: 
\begin{align*}
& U_\eta(t)=W_\eta(t)V_\eta'(t)\in\mathcal{M}_{(n+1)\times(n+1)}(\mathbb{R}),\\
& P_\eta(t)=Q_\eta(t)^\top K_\eta(t)\in\mathcal{M}_{(n+1)\times(n+1)}(\mathbb{R}),
\end{align*}
for each $1\leq\eta\leq h$ and $t\in\mathbb R_0^+$. We still refer to the matrix $U_\eta(t)$ as the value matrix since it plays a similar role. Similarly, we define the following functions:
\begin{align*}
\alpha_{ij}^\eta:\mathbb R_0^+\times(\mathcal E_W^n)^\ell\to\mathbb R,\quad  & \alpha_{ij}^\eta(t,\by)=\frac{1}{Z_i^\eta(t,\by)}\exp(\y_i^\top P_\eta(t)\y_j),\\
Z_i^\eta:\mathbb R_0^+\times(\mathcal E_W^n)^\ell\to\mathbb R,\quad & Z_i^\eta(t,\by)=D_\eta(t)_{ii}^{-1}=\sqrt{n+1}\sum_{j=1}^\ell\exp(\y_i^\top P_\eta(t)\y_j),
\end{align*}
respectively, for each $1\leq i,j\leq\ell$, $1\leq\eta\leq h$, $t\in\mathbb R_0^+$ and $\by=(\y_1,\hdots,\y_\ell)\in(\mathcal E_W^n)^\ell$. The matrix having $\alpha_{ij}^\eta$ as its $i$-th row and $j$-th column entry is usually called the attention matrix of head $\eta$.

With the notation just introduced, the dynamical system that describes the evolution of $\ell$ tokens on the ellipsoid $\mathcal E_W^n$ according to a transformer with $h$ heads is given by:
\begin{equation}\label{Transformer}
\boxed{\dot\y_i=T_{\y_i}\pi_W\cdot\left(\sum_{\eta=1}^h\sum_{j=1}^\ell\alpha_{ij}^\eta(t,\by)U_\eta(t)\y_j\right),
}
\end{equation}
for each $1\leq i\leq \ell$, $t\in\mathbb R_0^+$ and $\by=(\y_1,\hdots,\y_\ell)\in(\mathcal E_W^n)^\ell$. 


\section{Transformers as gradient vector fields}\label{sec:grad}

It was noted in~\cite{GeLePoRi2024b} that the transformer dynamics can be regarded as a gradient vector field under certain assumptions. For the benefit of the readers we formally prove such observation in the slightly more general setting where $P$ is not the identity matrix. Throughout this section, we consider the particular case of \cref{Transformer} described by the following assumptions.

\begin{hypothesis}\label{ass:grad}
There is only one head, $h=1$, $W=P$ and we have:
\begin{enumerate}
    \item $U_1(t)=\mathbb I_{n+1}$, and
    \item $P_1(t)=P$ is time-independent, positive definite, and symmetric.
\end{enumerate}
\end{hypothesis}

\subsection{Riemannian metric on the configuration space}
A Riemannian metric $g$ on $(\mathbb R_0^{n+1})^\ell$ may be defined as follows:
\begin{equation}\label{eq:metric}
\langle X_\bx,Y_\bx\rangle_{g(\bx)}=\sum_{i=1}^\ell Z_i(\bx)X_{\x_i}^\top P Y_{\x_i},
\end{equation}
for each $X_\bx=(X_{\x_1},\hdots,X_{\x_\ell}),Y_\bx=(Y_{\x_1},\hdots,Y_{\x_\ell})\in T_\bx(\mathbb R_0^{n+1})^\ell$ and $\bx=(\x_1,\hdots,\x_\ell)\in(\mathbb R_0^{n+1})^\ell$. For each $\by\in(\mathcal E_P^n)^\ell$, the orthogonal decomposition induced by $g$ 
is denoted by:
\begin{equation*}
T_\by(\mathbb R_0^{n+1})^\ell=T_\by(\mathcal E_P^n)^\ell\oplus T_\by^\perp(\mathcal E_P^n)^\ell,\quad X_\by =X_\by ^\parallel+X_\by ^\perp,
\end{equation*}
where $T^\perp(\mathcal E_P^n)^\ell\to(\mathcal E_P^n)^\ell$ denotes the normal bundle, i.e.: 
\begin{equation*}
T_\by ^\perp(\mathcal E_P^n)^\ell=\{X_\by\in T_\by(\mathbb R_0^{n+1})^\ell\mid \langle X_\by ,Y_\by\rangle_{g(\by)} =0,~\forall~Y_\by \in T_\by(\mathcal E_P^n)^\ell\}.
\end{equation*}
The orthogonal projection is the following vertical bundle morphism over $(\mathcal E_P^n)^\ell$:
\begin{equation*}
\boldsymbol\pi^\parallel:T(\mathbb R_0^{n+1})^\ell|_{(\mathcal E_P^n)^\ell}\to T(\mathcal E_P^n)^\ell,\quad X_\by\mapsto\boldsymbol\pi_\by^\parallel(X_\by)=X_\by^\parallel.
\end{equation*}

The following lemma gives the explicit expression of the orthogonal projection.

\begin{lemma}\label{lemma:orthogonalprojection}\rm
Under the previous conditions, the orthogonal projection is given by $\boldsymbol\pi^\parallel=T\boldsymbol\pi_P|_{(\mathcal E_P^n)^\ell}$.
\end{lemma}

\begin{proof}
It is enough to prove that, for each $\by=(\y_1,\hdots,\y_\ell)\in(\mathcal E_P^n)^\ell$ and $X_\by=(X_{\y_1},\hdots,X_{\y_\ell})\in T_\by(\mathbb R_0^{n+1})^\ell$, we have that $X_\by-T_\by\boldsymbol\pi_P\cdot X_\by\in T_\by^\top(\mathcal E_P^n)^\ell$, i.e., that $\langle X_\by-T_\by\boldsymbol\pi_P\cdot X_\by,Y_\by\rangle_{g(\by)}=0$ for each $Y_\by=(Y_{\y_1},\hdots,Y_{\y_\ell})\in T_\by(\mathcal E_P^n)^\ell$. By using \cref{eq:TpiW} and \cref{eq:metric}, this latter condition is clearly satisfied:
\begin{align*}
\langle X_\by-T_\by\boldsymbol\pi_P\cdot X_\by ,Y_\by \rangle_{g(\by)} & =\sum_{i=1}^\ell Z_i(\by)(X_{\y_i}-X_{\y_i}+\y_i^\top P X_{\y_i}\y_i)^\top P Y_{\y_i}\\
& =\sum_{i=1}^\ell Z_i(\by)\y_i^\top P X_{\y_i}\underbrace{\y_i^\top P Y_{\y_i}}_0=0,
\end{align*}
where we have used \cref{remark:tangentbundles}.
\end{proof}

Lastly, recall that $\imath_P:(\mathcal E_P^n)^\ell\hookrightarrow(\mathbb R_0^{n+1})^\ell$ is an embedding (and, in particular, an immersion). Hence, we can pullback $g$ to the Riemannian metric $g_P=\imath_P^*g$ on $(\mathcal E_P^n)^\ell$.

\subsection{Gradient vector field}

Let us show that the transformer dynamics is a gradient vector field on the manifold $(\mathcal E_P^n)^\ell$ equipped with the Riemannian metric $g_P=\imath_P^*g$. For simplicity, we introduce the following vector fields corresponding to \cref{Transformer} under \cref{ass:grad} (before and after projecting to the ellipsoid, respectively):
\begin{align*}
X_P:(\mathbb R_0^{n+1})^\ell\to T(\mathbb R_0^{n+1})^\ell,\quad & \bx\mapsto X_P(\bx)=\begin{pmatrix}
\sum_{j=1}^\ell\alpha_{1j}(\bx)\x_j\\
\vdots\\
\sum_{j=1}^\ell\alpha_{\ell j}(\bx)\x_j
\end{pmatrix},\\
Y_P:(\mathcal{E}_P^n)^\ell\to T(\mathcal{E}_P^n)^\ell,\quad & \by\mapsto Y_P(\by)=\begin{pmatrix}
T_{\y_1}\pi_P\cdot\left(\sum_{j=1}^\ell\alpha_{1j}(\by)\y_j\right)\\
\vdots\\
T_{\y_\ell}\pi_P\cdot\left(\sum_{j=1}^\ell\alpha_{\ell j}(\by)\y_j\right)
\end{pmatrix}.
\end{align*}
Note that $Y_P(\by)=T_\by\boldsymbol\pi_P\cdot X_P(\by)$. Now we show that $X_P$ is a gradient field with the metric $g$.

\begin{lemma}\label{lemma:gradV}\rm
We have ${\rm grad}_g\,V=-X_P$ for the following the potential function:
\begin{equation}\label{eq:V}
V:(\mathbb R_0^{n+1})^\ell\to\mathbb R,\quad \bx=(\x_1,\hdots,\x_\ell)\mapsto V(\bx)=-\frac{1}{2}\sum_{i,j=1}^\ell\exp(\x_i^\top P\x_j).
\end{equation}
 
\end{lemma}

\begin{proof}
For each $1\leq k\leq \ell$, we have:
\begin{align*}
\frac{\partial V(\bx)}{\partial\x_k} & =-\frac{1}{2}\sum_{i,j=1}^\ell\exp(\x_i^\top P\x_j)(\delta_{ik} P\x_j+\delta_{kj} P^\top \x_i)
=-\sum_{i=1}^\ell\exp(\x_k^\top P\x_i)P\x_i,
\end{align*}
where $\delta_{ij}$ denotes the Kronecker delta and we have used that $P$ is symmetric. Therefore:
\begin{align*}
\frac{\partial V(\bx)}{\partial\bx}=\begin{pmatrix}\frac{\partial V}{\partial\x_1}(\bx) \\ \vdots \\ \frac{\partial V}{\partial\x_\ell}(\bx)\end{pmatrix}=-\begin{pmatrix} \sum_{j=1}^\ell\exp(\x_1^\top P\x_j)P\x_j \\ \vdots \\ \sum_{j=1}^\ell\exp(\x_\ell^\top P\x_j)P\x_j\end{pmatrix}.
\end{align*}
From this, \cref{eq:gradcoordinates} and \cref{eq:metric}, we conclude:
\begin{align*}
{\rm grad}_g\,V(\bx) & =-\begin{pmatrix} Z_1^{-1}(\bx )P^{-1}  & \dots & 0\\\vdots & \ddots & \vdots\\
0 & \dots & Z_\ell^{-1}(\bx )P^{-1} 
\end{pmatrix}\begin{pmatrix}\sum_{j=1}^\ell\exp(\x_1^\top P\x_j)P\x_j \\ \vdots \\ \sum_{i=j}^\ell\exp(\x_\ell^\top P\x_j)P\x_j\end{pmatrix}\\
& =-X_P(\bx).
\end{align*}
\end{proof}

The previous result, together with the fact that the gradient on a submanifold of a Riemannian manifold is the orthogonal projection of the gradient on the original manifold, enable us to show that $Y_P$ is a gradient vector field.

\begin{theorem}\label{theorem:gradientVN}\rm
Let $V_P=V\circ\imath_P:(\mathcal E_P^n)^\ell\to\mathbb R$, then ${\rm grad}_{g_P}\,V_P=-Y_P$. 
\end{theorem}

\begin{proof}
For each $Z\in \mathfrak X\left((\mathcal E_P^n)^\ell\right)$ and $\by\in(\mathcal E_P^n)^\ell$, we have:
\begin{align*}
\mathbf{d}_\by V_P(Z(\by)) & =\mathbf{d}_{\by}(V\circ\imath_P)(Z(\by))\\
& =\mathbf{d}_{\imath_P(\by)}V(T_\by\imath_P\cdot Z(\by))\\
& =\langle ({\rm grad}_g\,V )(\imath_P(\by)), T_y \imath_P\cdot Z(\by) \rangle_{g(\imath_P(\by))}\\
& =\langle ({\rm grad}_g\,V )^{\parallel}(\imath_P(\by))+({\rm grad}_g\,V )^{\perp}(\imath_P(\by)), T_\by \imath_P\cdot Z(\by) \rangle_{g(\imath_P(\by))}\\
& =\langle ({\rm grad}_g\,V )^{\parallel}(\imath_P(\by)), T_{\by} \imath_P\cdot Z(\by) \rangle_{g(\imath_P(\by))}\\
& =\langle T_{\imath_P(\by)}\boldsymbol\pi_P\cdot ({\rm grad}_g\,V )(\imath_P(\by)), T_\by \imath_P\cdot Z(\by) \rangle_{g(\imath_P(\by))}\\
& =\langle -T_\by\imath_P\cdot Y_P(\by), T_\by \imath_P\cdot Z(\by) \rangle_{g(\imath_P(\by))}\\
& =\langle - Y_P(\by),  Z(\by) \rangle_{h(\by)},
\end{align*}
where we used \cref{lemma:orthogonalprojection} and the equality $T_\by\imath_P\cdot Y_P(\by)=Y_P(\by)$, which follows from regarding $Y_P(\by)$ both as an element of $T_\by(\mathbb R_0^{n+1})^\ell$ and $T_\by(\mathcal E_P^n)^\ell$.
\end{proof}

\subsection{Stability analysis}

Having established that \cref{Transformer}, under \cref{ass:grad}, is a gradient vector field, it is natural to use the potential $V_P:(\mathcal E_P^n)^\ell\to\mathbb R$ as a Lyapunov function to study the asymptotic behavior of the tokens. In order to stablish that all trajectories converge to an equilibrium, we first prove that they converge to the zeroes of the gradient field.

\begin{lemma}\label{lemma:lasalle}\rm
The trajectories of \cref{Transformer} under \cref{ass:grad} converge to the set:
\begin{equation*}
\{\by\in(\mathcal E_P^n)^\ell\mid{\rm grad}_{g_P}\,V_P(\by)=0\}.
\end{equation*}
\end{lemma}

\begin{proof}
Let $\by\in(\mathcal E_P^n)^\ell$ and $Y_\by=(Y_{\y_1}, \hdots,Y_{\y_\ell})\in T_\by(\mathcal E_P^n)^\ell$. Recall that the formal time derivative (at $t=0$) of the potential $V_P$ is the map $\dot V_P=\mathbf{d}V_P(Y_P):(\mathcal E_P^n)^\ell\to\mathbb R$. From \cref{theorem:gradientVN}, we obtain:
\begin{align*}
\dot V_N=\mathbf{d}V_P(Y_P)=\mathbf{d}V_P(-{\rm grad}_{g_P}\,V_P)=-\langle{\rm grad}_{g_P}\,V_P,{\rm grad}_{g_P}\,V_P\rangle_h\leq0,
\end{align*}
and the equality holds if and only if ${\rm grad}_{g_P}\,V_P=0$. The proof is concluded by a routine application of LaSalle's invariance principle.
\end{proof}

\begin{theorem}\rm\label{theorem:grad}
If \cref{ass:grad} holds, then every trajectory of \cref{Transformer} converges to an equilibrium.
\end{theorem}

\begin{proof}
Recall that the potential $V_P$ satisfies the Łojasiewicz inequality if $|V_P|\leq\lambda|{\rm grad}_{g_P}\,V_P|_h$ for some $\lambda>0$. A sufficient condition for the Łojasiewicz inequality to hold is that $((\mathcal E_P^n)^\ell,g_P=\imath^*g)$ is a real analytic Riemannian manifold and the potential is real analytic, i.e., $V_P\in C^\omega((\mathcal E_P^n)^\ell)$. It is clear that these two conditions are satisfied since $(\mathcal E_P^n)^\ell$ is a real analytic submanifold of $\mathbb R^{n+1}$ and $Z_i,\alpha_{ij}\in C^\omega((\mathbb R_0^{n+1})^\ell),\mathbb R^+)$ for each $1\leq i,j\leq\ell$, which ensures that both the Riemannian metric $g_P$ and the potential $V_P$ are real analytic.

On the other hand, $(\mathcal E_P^n)^\ell\subset(\mathbb R_0^{n+1})^\ell$ is compact, whence the set of $\omega$-limit points of \cref{Transformer} is non-empty. The Łojasiewicz inequality thus ensures that every trajectory converges to a point $\by\in(\mathcal E_P^n)^\ell$. From \cref{lemma:lasalle}, we know that $\by$ is an equilibrium of \cref{Transformer} since $\dot\by=-{\rm grad}_{g_P}\,V_P(\by)=0$.
\end{proof}

If we take $P$ to be the identity, linearization of $Y_P$ around each equilibrium point shows that the only equilibria that are asymptotically stable are the consensus equilibria, i.e., the points $\by=(\y_1,\hdots,\y_\ell)\in(\mathbb S^n)^\ell$ satisfying $\y_i=\y_j$ for every $i,j\in\{1,2,\hdots,\ell\}$. This linearization strategy was employed, e.g., in~\cite{markdahl2017almost}. Unfortunately, when $P$ is not the identity this strategy leads to conditions whose validity cannot be easily ascertained.

\section{Full self-attention matrix}\label{sec:full}
In this section we consider a particular case of the model \cref{Transformer} described by the following assumptions.

\begin{hypothesis}\label{ass:hemisphere}
For each head $1\leq\eta\leq h$, we have:
\begin{enumerate}
    \item $U_\eta(t)=\mathbb I_{n+1}$,
    \item $P_\eta(t)$ is bounded, i.e., $\sup_{t\in\mathbb R_0^+}\|P_\eta(t)\|<\infty$, and
    \item $P_\eta(t)$ is uniformly continuous on $\mathbb R_0^+$.
\end{enumerate}
\end{hypothesis}

In the proof of \cref{theorem:hemisphere} below, a nonsmooth candidate for Lyapunov function will be introduced. In order to handle this situation, we briefly recall how to compute the Dini derivative of a function defined through a maximum (cf. \S2.3 of~\cite{LiFrMa2007}).

\begin{definition}\rm
The \emph{upper Dini derivative} of a continuous function $f:\mathbb R\to\mathbb R$ is defined as:
\begin{equation*}
\dot f^+(t)=\lim\sup_{\tau\to t^+}\frac{f(t+\tau)-f(t)}{\tau},\qquad t\in\mathbb R.
\end{equation*}
\end{definition}

As a particular case, let $\{f_i:\mathbb R_0^+\to\mathbb R\mid i\in I\}$ be a family of continuously differentiable functions, and consider its maximum:
\begin{equation*}
f:\mathbb R_0^+\to\mathbb R,\quad t\mapsto f(t)=\max_{i\in I}f_i(t).
\end{equation*}
For each $t\in\mathbb R_0^+$, the upper Dini derivative of $f$ is computed according to Danskin's theorem (cf. Lemma~2.2 of~\cite{LiFrMa2007}):
\begin{equation}\label{eq:Dini_max}
\dot f^+(t)=\max_{i\in\mathcal I(t)}\dot f_i(t),
\end{equation}
where $\mathcal I(t)=\{i\in I\mid f(t)=f_i(t)\}$. In addition, let us prove the following two lemmas.

\begin{lemma}\rm\label{lemma:aux1}
If there exists $b>0$ such that $\max_{1\leq\eta\leq h}\sup_{t\in\mathbb R_0^+}\|P_\eta(t)\|\leq b$, then there exist $c_1,c_2>0$ such that $c_1\leq\alpha_{ij}^\eta(t,\by)\leq c_2$ for each $t\in\mathbb R_0^+$, $\by\in(\mathcal E_W^n)^\ell$, $1\leq\eta\leq h$ and $1\leq i,j\leq\ell$. 
\end{lemma}

\begin{proof}
By compactness of $\mathcal E_W^n $, there exists $K>0$ such that $|\y|\leq K$ for each $\y\in\mathcal E_W^n $. Therefore, we have $|\y_i^\top P_\eta(t)\y_j|\leq K^2\|P_\eta(t)\|\leq K^2b$ and, thus, $\exp(-K^2 b)\leq\exp(\y_i^\top P_\eta(t)\y_j)\leq\exp(K^2 b)$. Moreover, we also have:
$$\exp(-K^2 b)\leq\sum_{k=1}^\ell\exp(\y_i^\top P_\eta(t)\y_k)\leq \ell\exp(K^2 b),$$
which leads to:
\begin{align*}
\alpha_{ij}^\eta(t,\by) & =\frac{\exp(\y_i^\top P_\eta(t)\y_j)}{\sqrt{n+1}\sum_{k=1}^i\exp(\y_i^\top P_\eta(t)\y_k)}\\
& \geq\frac{\exp(-K^2 b)}{\sqrt{n+1}\ell\exp(K^2 b)}\geq\frac{1}{\sqrt{n+1}\ell\exp(2 K^2 b)}.
\end{align*}
Similarly, we have:
\begin{equation*}
\alpha_{ij}^\eta(t,\by)=\frac{\exp(\y_i^\top P_\eta(t)\y_j)}{\sqrt{n+1}\sum_{k=1}^i\exp(\y_i^\top P_\eta(t)\y_k)}\leq\frac{\exp(K^2 b)}{\exp(-K^2 b)}=\exp(2 K^2 b).
\end{equation*}
By taking $c_1=1/(\sqrt{n+1}\ell\exp(2 K^2 b))$ and $c_2=\exp(2 K^2 b)$, we conclude.
\end{proof}

\begin{lemma}\label{lemma:unifcont}\rm
Let $\by=(\y_1,\hdots,\y_\ell):\mathbb R_0^+\to\mathcal E_W^n $ be a solution of \cref{Transformer} under \cref{ass:hemisphere}, and $v\in\mathcal E_W^n $. If there exists $b>0$ such that $\sup_{t\in\mathbb R_0^+}\|P_\eta(t)\|\leq b$ and $P_\eta$ is uniformly continuous on $\mathbb R_0^+$ for each $1\leq\eta\leq h$, then the following functions:
\begin{equation*}
f_i:\mathbb R_0^+\to\mathbb R,\quad t\mapsto f_i(t)=\sum_{\eta=1}^h\sum_{j=1}^\ell\alpha_{ij}^\eta(t,\by(t))\left(v^\top\y_j(t)-\y_i(t)^\top W\y_j(t)v^\top\y_i(t)\right),
\end{equation*}
for each $1\leq i\leq\ell$, are bounded and uniformly continuous.
\end{lemma}

\begin{proof}
Let $1\leq i\leq\ell$. It is clear that the following functions:
\begin{align*}
g_j:\mathbb R_0^+\to\mathbb R,\quad & t\mapsto g_j(t)=v^\top\y_j(t)-\y_i(t)^\top W\y_j(t)v^\top\y_i(t),\\
g_j^\eta:\mathbb R_0^+\to\mathbb R,\quad & t\mapsto g_j^\eta(t)=\alpha_{ij}^\eta(t,\by(t)),
\end{align*}
for each $1\leq\eta\leq h$ and $1\leq j\leq\ell$, are bounded: $g_j$ due to tokens evolving on the ellipsoid, and $g_j^\eta$ due to \cref{lemma:aux1}. Recall that the addition and multiplication of bounded and uniformly continuous functions results in uniformly continuous functions. Thus, it is enough to prove that $g_j$ and $g_j^\eta$ are uniformly continuous to conclude that $f_i$ is uniformly continuous:
\begin{enumerate}
    \item The derivative of $g_j$ is bounded on $\mathbb R_0^+$ since the tokens evolve on the ellipsoid and their dynamics is given by \cref{Transformer}. Note that $\sup_{t\in\mathbb R_0^+}|\dot\y_j(t)|<\infty$ for each $1\leq j\leq\ell$ thanks to \cref{lemma:aux1}. Hence, $g_j$ is uniformly continuous on $\mathbb R_0^+$.
    \item Given that the tokens evolve on the ellipsoid and every $P_\eta$ is bounded on $\mathbb R_0^+$, we can ensure the existence of $K>0$ such that:
    $$\max_{1\leq\eta\leq h}\sup_{t\in\mathbb R_0^+}|\y_i(t)^\top P_\eta(t)\y_j(t)|\leq K.$$
    Moreover, $\exp|_{[-K,K]}:[-K,K]\to[\exp(-K),\exp(K)]$ is uniformly continuous, as it is defined on a compact. Hence, $\exp(\y_i(\cdot)P_\eta(\cdot)\y_j(\cdot)):\mathbb R_0^+\to[\exp(-K),\exp(K)]$ is uniformly continuous, as the composition of uniformly continuous functions is uniformly continuous. In particular, $Z_i^\eta(\cdot,\by(\cdot)):\mathbb R_0^+\to[\ell\exp(-K),\ell\exp(K)]$ is uniformly continuous. By gathering all, we conclude that $g_j^\eta$ is uniformly continuous on $\mathbb R_0^+$.
\end{enumerate}
\end{proof}

The next result claims attractivity of the consensus set provided that the initial position of the tokens is in some open hemisphere of the ellipsoid.

\begin{theorem}\label{theorem:hemisphere}\rm
Let $v\in\mathcal E_W^n $ and consider the open hemisphere:
\begin{equation*}
\mathcal H^+(v)=\{\y\in\mathcal E_W^n \mid v^\top y>0\}.
\end{equation*}
If \cref{ass:hemisphere} holds, then the consensus set $\mathcal C_\ell^+(v)$ in the product of hemispheres $\mathcal H^+(v)^\ell$, given by:
\begin{equation*}
\mathcal C_\ell^+(v)=\{\by=(\y,\hdots,\y)\in(\mathcal E_W^n)^\ell\mid\y\in\mathcal H^+(v)\},
\end{equation*}
is attractive for \cref{Transformer} with domain of attraction $\mathcal H^+(v)^\ell$.
\end{theorem}

\begin{proof}
For each $t\in\mathbb R_0^+$ and $\by=(\y_1,\hdots,\y_\ell)\in\mathcal H^+(v)^\ell$, note that:
\begin{equation*}
v^\top \sum_{\eta=1}^h\sum_{j=1}^\ell\alpha_{ij}^\eta(t,\by)\y_j=\sum_{\eta=1}^h\sum_{j=1}^\ell\alpha_{ij}^\eta(t,\by)v^\top\y_j>0,
\end{equation*}
since $\alpha_{ij}^\eta(t,\by)>0$ for each $1\leq i,j\leq\ell$ and $1\leq\eta\leq h$. Therefore, $\dot\y_i$ points to the interior of $\mathcal H^+(v)$ for each $1\leq i\leq\ell$, which ensures that $\mathcal H^+(v)^\ell$ is forward invariant for \cref{Transformer} under \cref{ass:hemisphere}.

On the other hand, an easy check shows that every $\by\in\mathcal C_\ell^+(v)$ is an equilibrium of \cref{Transformer} under \cref{ass:hemisphere}. Let us define the following function:
\begin{equation}\label{LyapunovFunction}
V:\mathcal H^+(v)^\ell\to\mathbb R,\quad\by\mapsto V(\by)=\max_{1\leq i\leq\ell}V_i(\by),
\end{equation}
where $V_i(\by)=1-v^\top\y_i$. Let $\by=(\y_1,\hdots,\y_\ell):\mathbb R_0^+\to(\mathcal E_W^n)^\ell$ be a solution of \cref{theorem:hemisphere} with $\by(0)\in\mathcal H^+(v)^\ell$. Forward invariance ensures that $\by(t)\in\mathcal H^+(v)^\ell$ for each $t\in\mathbb R_0^+$. Moreover, let $\mathcal I(t)=\{i\in\{1,\hdots,\ell\}\mid V(\by(t))=V_i(\by(t))\}$. Note that, for $i\in\mathcal I(t)$, we have that $v^\top\y_i\leq v^\top\y_j$ for each $1\leq j\leq\ell$, where we dropped the argument $t$ for simplicity. Given that $\y_i^\top W\y_j\leq 1$ (since all tokens lie on the ellipsoid) and $v^\top\y_i>0$ (by assumption), we get:
\begin{equation*}
\y_i^\top W\y_j v^\top\y_i\leq v^\top\y_i\leq v^\top\y_j,\qquad 1\leq j\leq\ell,~i\in\mathcal I(t).
\end{equation*}
From this and \cref{eq:Dini_max}, the upper Dini derivative of $V(\mathbf y(t))$, $t\in\mathbb R_0^+$, is given by:
\begin{align*}
\dot V^+(t,\by(t)) & =\max_{i\in\mathcal I(t)}\dot V_i(t,\by(t))=-\min_{i\in\mathcal I(t)}v^\top \dot\y_i(t)\\
& =-\min_{i\in\mathcal I(t)}\sum_{\eta=1}^h\sum_{j=1}^\ell\alpha_{ij}^\eta(t,\by)\left(v^\top\y_j(t)-\y_i(t)^\top W\y_j(t)v^\top\y_i(t)\right)\leq 0.
\end{align*}
The equality holds if and only if $v^\top\y_i(t)=v^\top\y_j(t)$ and $\y_i(t)^\top W\y_j(t)=1$ for each $1\leq j\leq\ell$, i.e., if and only if $\by(t)\in\mathcal C_\ell^+(v)$. In addition, the minimum of bounded and uniformly continuous functions is bounded and uniformly continuous. Hence, \cref{lemma:unifcont} ensures that $\dot V^+$ is uniformly continuous on $\mathbb R_0^+$. Therefore, $V$ is a strict Lyapunov function for $\mathcal C_\ell^+(v)$ on $\mathcal H^+(v)^\ell$, and we conclude by the Lyapunov-like theorem based on Barbalat's lemma (cf. Theorem~8.4 of~\cite{khalil2002nonlinear}).
\end{proof}


\begin{remark}[Autonomous systems]
For the case where the matrices $\{P_\eta\mid 1\leq\eta\leq h\}$ are time-independent, attractivity to the consensus set in the previous theorem can be upgraded to asymptotic stability by applying Lasalle's invariance principle instead of Barbalat's lemma.
\end{remark}

\begin{remark}[Closest result available in the literature] Similar conclusions appear in~\cite{GeLePoRi2024a} (see Lemma 4.2) under the stronger assumptions of a single attention head and that both $U$ and $P=Q^\top K$ are the identity matrix.
\end{remark}

\begin{remark}[Higher dimensions]
Let us restrict ourselves to the case where we have normalization to the sphere, i.e., $W=\mathbb I_{n+1}$. Wendel's theorem (cf. Eq. (1) of~\cite{We1962}) gives the probability that $\ell$ tokens lie on the same hemisphere when distributed uniformly at random; namely:
\begin{equation*}
\mathcal P_{\ell,n}=\frac{1}{2^{\ell-1}}\sum_{\mu=0}^{n-1}\binom{\ell-1}{\mu}.
\end{equation*}
In particular, $\mathcal P_{\ell,n}=1$ whenever $n\geq\ell$. As a result, if the starting position of the tokens is chosen from a uniformly random distribution and $n\geq\ell$, then they will lie on the same hemisphere almost surely. The previous result thus deals with the most general situation for higher dimensions.
\end{remark}

\section{Auto-regressive self-attention matrix}\label{sec:autoregressive}

This section addresses the auto-regressive (also known as causal) case, that is, the case where the dynamics of each token only depends on itself and the previous tokens. This corresponds to the model \cref{Transformer} with the so-called \emph{auto-regressive self-attention matrix}, i.e.:
\begin{align*}
& \alpha_{ij}^\eta(t,\by)=\left\{\begin{array}{ll}
\displaystyle\frac{1}{Z_i^\eta(t,\by)}\exp(\y_i^\top P_\eta(t)\y_j), & i\geq j,\\
0, & i<j,
\end{array}\right.\\
& Z_i^\eta(t,\by)=\sqrt{n+1}\sum_{j=1}^i\exp(\y_i^\top P_\eta(t)\y_j).
\end{align*}
Note that the equations are decoupled and, thus, the solution of the $i$-th equation only depends on the first $i$-th initial conditions. Hence, given an initial condition $\by^0=(\y_1^0,\hdots,\y_\ell^0)\in(\mathcal E_W^n)^\ell$, we denote the solution of the $i$-th equation by $\y_i(\cdot,\y_1^0,\hdots,\y_i^0):\mathbb R_0^+\to\mathcal E_W^n $, and the solution of the system by:
\begin{equation*}
\by(\cdot,\by^0)=(\y_1(\cdot,\y_1^0),\hdots,\y_\ell^0(\cdot,\y_1^0,\hdots,\y_\ell^0)):\mathbb R_0^+\to(\mathcal E_W^n)^\ell.
\end{equation*}
For later convenience, let us introduce the following functions:
\begin{equation}\label{eq:tildealpha}
\tilde\alpha_{ij}^\eta:\mathbb R_0^+\to\mathbb R_0^+,\quad t\mapsto\tilde\alpha_{ij}^\eta(t)=\alpha_{ij}^\eta(t,\by(t,\by^0)),
\end{equation}
for each $1\leq\eta\leq h$ and $1\leq i,j\leq\ell$.

\subsection{Identity value matrix}\label{sec:masked_id}

Let us consider the case where $W=\mathbb I_{n+1}$, i.e., the tokens evolve on the sphere.

\begin{hypothesis}\label{ass:consensussystem}
The model \cref{Transformer} is auto-regressive, \mbox{$W=\mathbb I_{n+1}$} and, for each head $1\leq\eta\leq h$, we have:
\begin{enumerate}
    \item $U_\eta(t)=\mathbb I_{n+1}$, and
    \item $P_\eta(t)$ is bounded, i.e., $\sup_{t\in\mathbb R_0^+}\|P_\eta(t)\|<\infty$.
\end{enumerate}
\end{hypothesis}


It is straightforward that, under the previous conditions, $\dot\y_1=0$ and, thus, the first token remains fixed: \mbox{$\y_1(t)=\y_1^0$} for each $t\in\mathbb R_0^+$. For each other token $\y_i$, $2\leq i\leq\ell$, the inner product between $\y_1^0$ and $\y_i$, i.e., the (cosine of the) angle between them, provides a projection of the dynamics onto the real line. In other words, using this angle we construct a scalar differential equation governing the evolution of the projection of the token on the real line. It them becomes simple to construct a Lyapunov function for the second token, ensuring convergence to $\y_1^0$ for every initial condition except for $\y_2(0)=-\y_1^0$. For the remaining tokens, an input-to-state stability argument coupled with the triangular nature of the dynamics leads to the asymptotic stability of the consensus set for almost all initial conditions. More specifically, let us start by proving the following lemma.

\begin{lemma}\label{lemma:aux2}\rm
Consider a point $\y^0\in\mathbb S^n$ and let $\tilde\alpha:\mathbb R_0^+\to[c,\infty[$ be a continuously differentiable function for some $c\in\mathbb R^+$. The only equilibria $\y^*\in\mathbb S^n$ of the following  differential equation:
\begin{equation}\label{eq:aux2}
\dot\y=\tilde\alpha(t)(\y^0-\y^\top\y^0\y),\qquad t\in\mathbb R_0^+,~\y\in\mathbb S^n,
\end{equation}
are $\y^*=-\y^0$ and $\y^*=\y^0$. Furthermore, the former is unstable whereas the latter is asymptotically stable with domain of attraction $\mathbb S^n-\{-\y^0\}$.

\end{lemma}

\begin{proof}
For the first part, note that the equation $\tilde\alpha(t)(\y^0-(\y^*)^\top\y^0\y^*)=0$ holds for each $t\in\mathbb R_0^+$ if and only if $(\y^*)^\top\y^0\in\{-1,1\}$, where we have used that $\tilde\alpha$ is always positive and both $\y^0$ and $\y^*$ lie on the $n$-sphere, i.e., if and only if $\y^*=-\y^0$ or $\y^*=\y^0$. For the second part, let $\y:\mathbb R_0^+\to\mathbb S^n$ be a solution of \cref{eq:aux2} and define:
\begin{equation*}
a:\mathbb R_0^+\to[-1,1],\quad t\mapsto a(t)=(\y^0)^\top\y(t).
\end{equation*}
From \cref{eq:aux2}, the dynamics of $a$ is easily seen to be given by:
\begin{equation*}
\dot a=\tilde\alpha(t)(\y^0)^\top (\y^0-\y^\top\y^0\y)=\tilde\alpha(t)\left(1-\left((\y^0)^\top\y\right)^2\right)=\tilde\alpha(t)(1-a^2).
\end{equation*}
The equilibria of the previous ODE are $a^*=-1$ and $a^*=1$, which correspond to $\y^*=-\y^0$ and $\y^*=\y^0$, respectively. To check that the former is unstable and the latter is asymptotically stable with domain of attraction $]-1,1]$, which corresponds to $\mathbb S^n-\{-\y^0\}$, we define:
\begin{equation}\label{eq:lyapunov}
V:[-1,1]\to\mathbb R,\qquad a\mapsto V(a)=\frac{2}{3}-a+\frac{a^3}{3}.
\end{equation}
We have that $V(a)\in\mathbb R_0^+$ for $-1\leq a\leq 1$ and $V(a)=0$ if and only if $a=1$. Moreover, its derivative is given by $V'(a)=a^2-1$, whence:
\begin{align*}
\dot V(t,a) & =V'(a)\dot a=-\tilde\alpha(t)V'(a)^2\leq -c V'(a)^2.
\end{align*}
Since $V'(a)\neq 0$ for each $-1<a<1$, we conclude that $V$ is a Lyapunov function for the equilibrium $a^*=1$ and its domain of attraction is $]-1,1]$.
\end{proof}

\begin{theorem}\rm\label{theorem:consensussytem}
If \cref{ass:consensussystem} holds, then the consensus set:
\begin{equation*}
\mathcal C_\ell=\{\by=(\y,\hdots,\y)\in(\mathbb S^n)^\ell\},
\end{equation*}
is asymptotically stable for the system \cref{Transformer} and the domain of attraction contains the following set:
\begin{equation*}
\mathcal D_\ell^1=\{(\y_1,\hdots,\y_\ell)\in(\mathbb S^n)^\ell\mid\y_j\neq-\y_1,~2\leq j\leq\ell\}.
\end{equation*}
\end{theorem}

\begin{proof}
Let $\by^0=(\y_1^0,\hdots,\y_\ell^0)\in(\mathbb S^n)^\ell$. To begin with, note that $\dot\y_1=0$ and, thus, the solution of the first equation is constant, i.e., $\y_1(t,\y_1^0)=\y_1^0$ for each $t\in\mathbb R_0^+$. By substituting this into the second equation, we may write $\dot\y_2=\sum_{\eta=1}^h\tilde\alpha_{21}^\eta(t)(\y_1^0-\y_2^\top\y_1^0\y_2)$. From \cref{lemma:aux1,lemma:aux2}, we conclude that $\y_2^*=\y_1^0$ is the only asymptotically stable equilibrium with domain of attraction $\mathbb S^n-\{-\y_1^0\}$. In particular, $\mathcal C_2$ is asymptotically stable for the subsystem of \cref{Transformer} under \cref{ass:consensussystem} given by the first two tokens, and the domain of attraction is $\mathcal D_2^1$.

We proceed by induction: given $2\leq i\leq\ell$, for each $2\leq j\leq i-1$ suppose that $\mathcal C_j$ is asymptotically stable for the subsystem of \cref{Transformer} under \cref{ass:consensussystem} given by the first $j$ tokens, and the domain of attraction contains $\mathcal D_j^1$. 

In order to study the behavior of the $i$-th token, we define the errors as:
\begin{equation*}
e_j:\mathbb R_0^+\to\mathbb R^{n+1},\quad t\mapsto e_j(t)=\y_j(t,\y_1^0,\dots,\y_j^0)-\y_1^0,\qquad 1\leq j\leq i-1.
\end{equation*}
Although $e_1=0$, it will be convenient to consider $e=(e_1,\hdots,e_{i-1})$. The dynamics of the $i$-th token may be written as:
\begin{align}\nonumber
\dot\y_i & =\sum_{\eta=1}^h\tilde\alpha_{i1}^\eta(t)(\y_1^0-\y_i^\top\y_1^0\y_i)+\sum_{\eta=1}^h\sum_{j=2}^{i-1}\tilde\alpha_{ij}^\eta(t)(\y_1^0+e_j-\y_i^\top\y_1^0\y_i-\y_i^\top e_j\y_i)\\\label{eq:dotyi}
& =\sum_{\eta=1}^h\sum_{j=1}^{i-1}\tilde\alpha_{ij}^\eta(t)(\y_1^0-\y_i^\top\y_1^0\y_i)+\sum_{\eta=1}^h\sum_{j=2}^{i-1}\tilde\alpha_{ij}^\eta(t)(e_j-\y_i^\top e_j\y_i).
\end{align}
Analogous to the proof of \cref{lemma:aux2}, we define:
\begin{equation*}
a_i:\mathbb R_0^+\to[-1,1],\quad t\mapsto a_i(t)=(\y_1^0)^\top\y_i(t,\y_1^0,\hdots,\y_i^0).
\end{equation*}
Its dynamics is readily obtained from \cref{eq:dotyi}:
\begin{equation}\label{eq:dota}
\dot a_i=\sum_{\eta=1}^h\sum_{j=1}^{i-1}\tilde\alpha_{ij}^\eta(t)(1-a_i^2)+\sum_{j=2}^{i-1}\tilde\alpha_{ij}(t)e_j^\top (\y_1^0-\y_i a_i).
\end{equation}
Note that $a_i^*=-1$ and $a_i^*=1$ are the only equilibria of the previous ODE when $e=(e_2,\hdots,e_{i-1})=0$. Let us consider the function $V$ introduced in \cref{eq:lyapunov}. For the dynamics \cref{eq:dota}, it satisfies:
\begin{align*}
\dot V(t,a_i) 
& =-\sum_{\eta=1}^h\sum_{j=1}^{i-1}\tilde\alpha_{ij}^\eta(t)(1-a_i^2)^2+(a_i^2-1)\sum_{\eta=1}^h\sum_{j=2}^{i-1}\tilde\alpha_{ij}^\eta(t)e_j^\top (\y_1^0-\y_i a_i)\\
& \leq-\sum_{\eta=1}^h\sum_{j=1}^{i-1}\tilde\alpha_{ij}^\eta(t)V'(a_i)^2+(1-a_i^2)\sum_{\eta=1}^h\sum_{j=2}^{i-1}\tilde\alpha_{ij}^\eta(t)|e_j|(|\y_1^0|+|\y_i|\,|a_i|)\\
& \leq -c_1 h(i-1)V'(a_i)^2+2 c_2 h\sum_{j=2}^{i-1}|e_j|,
\end{align*}
where we have used that there exist $c_1,c_2>0$ such that $c_1\leq\tilde\alpha_{ij}^\eta(t)\leq c_2$ for each $t\in\mathbb R_0^+$ (recall \cref{lemma:aux1}). Note that $V'(1)=0$ and $V'|_{]-1,1[}\neq0$. As a result, the previous inequality, together with the fact that $V|_{]-1,1[}>0$ and $V(1)=0$, ensures that $V|_{]-1,1]}$ is an ISS-Lyapunov function for the equilibrium $a_i^*=1$ of \cref{eq:dota} where the input is given by $e=(e_1,\hdots,e_{i-1})$. We conclude that $a_i^*=-1$ is an unstable equilibrium whereas $a_i^*=1$ is ISS-stable on $]-1,1]$. For the system \cref{eq:dotyi}, this corresponds to $\y_i^*=-\y_1^0$ being unstable and $\y_i^*=\y_1^0$ being ISS-stable on $\mathbb S^n-\{-\y_1^0\}$. 

Lastly, if we regard the errors as functions of the initial conditions, i.e., $e(t)=e(t,\y_1^0,\hdots,\y_{i-1}^0)$, then the induction hypothesis ensures that $e^*=0$ is an asymptotically stable equilibrium and its domain of attraction contains $\mathcal D_{i-1}^1$. As a result, $(e^*,\y_i^*)=(0,\y_1^0)$ is asymptotically stable for the cascade system $(e,\y_i)$ and its domain of attraction contains $\mathcal D_i^1$ (cf. Lemma~4.7 of~\cite{Kh2002}).
\end{proof}

\begin{remark}[Closest result available in the literature]
Similar conclusions are reported under Theorem 4.1 in~\cite{KaPoRi2024} by imposing stronger assumptions, time invariance of $P=Q^\top K$  and existence of a single attention head, although the authors state that time-invariance is not explicitly used.
\end{remark}

\begin{remark}[Invertible value matrix with different choice of projection]\label{rem:masked_special_projection}
The results in this section can be applied to non-identity value matrices, i.e., $U\neq\mathbb I_{n+1}$. To that end, we need to substitute the projection $T\pi_W$ introduced in \cref{eq:TpiW} by a different projection to the ellipsoid. More specifically, we restrict ourselves to the single-head case, $h=1$, assume that $U$ is invertible and pick $W=U^\top U$, which is symmetric and positive-define by construction. Then, for each $\y\in\mathcal E_W^n $, we define a new projection as:
\begin{equation*}
(\Pi_W)_\y:T_\y\mathbb R_0^{n+1}\to T_\y\mathcal E_W^n ,\quad X_\y\mapsto(\Pi_W)_\y\cdot X_\y=U^{-1}(\mathbb I_{n+1}-U\y\y^\top U^\top)\cdot X_\y.
\end{equation*}
With these choices, we obtain the system:
\begin{equation}\label{Transformer_masked_specialPi}
\dot\y_i=(\Pi_W)_{\y_i}\cdot\left(\sum_{j=1}^i\alpha_{ij}(t,\by)U\y_j\right)=\sum_{j=1}^i\alpha_{ij}(t,\by)\left(\y_j-\y_i^\top W\y_j\y_i\right),
\end{equation}
for each $1\leq i\leq\ell$, $t\in\mathbb R_0^+$ and $\by=(\y_1,\hdots,\y_\ell)\in(\mathcal E_W^n)^\ell$. 

The change of coordinates $\z_i=U\y_i\in\mathbb S^n$ for each $1\leq i\leq\ell$ brings~\cref{Transformer_masked_specialPi} into:
\begin{equation*}
\dot\z_i=T_{\y_i}\pi\cdot\left(\sum_{j=1}^i\beta_{ij}(t,\bz)\z_j\right)=\sum_{j=1}^i\beta_{ij}(t,\bz)(\z_j-\z_i^\top \z_j\z_i),
\end{equation*}
for each $1\leq i\leq\ell$, $t\in\mathbb R_0^+$ and $\bz=(\z_1,\hdots,\z_\ell)\in\mathbb S^n$, where $\beta_{ij}(t,\bz)=\alpha_{ij}(t,U^{-1}\bz)$, $1\leq i,j\leq\ell$. In other words, we obtain \cref{Transformer} under \cref{ass:consensussystem} with $h=1$ and $\beta_{ij}$ instead of $\alpha_{ij}$, which also satisfy \cref{lemma:aux1}. Therefore, \cref{theorem:consensussytem} ensures that the consensus set $\mathcal C_\ell$ is asymptotically stable and the domain of attraction contains the set $\mathcal D_\ell^1$.
\end{remark}

\subsection{Symmetric value matrix}\label{sec:masked_symmetric}

Now we extend the results of the previous section to more general value matrices. As above, the tokens evolve on the sphere, i.e., $W=\mathbb I_{n+1}$.

\begin{hypothesis}\label{ass:consensyssystemU}
The model \cref{Transformer} is auto-regressive, \mbox{$W=\mathbb I_{n+1}$}, and we have:
\begin{enumerate}
    \item There is only one head, i.e., $h=1$.
    \item $U_1(t)=U$ with $U^\top=U$.
    \item $P_1(t)=P(t)$ is bounded, i.e., $\sup_{t\in\mathbb R_0^+}\|P(t)\|<\infty$.
\end{enumerate}
\end{hypothesis}


We denote the spectrum of $U$ by $\lambda(U)$. Note that $\lambda(U)\subset\mathbb R$ as $U$ is symmetric. Given $\lambda\in\lambda(U)$, the corresponding eigenspace is denoted by $L_\lambda(U)\subset\mathbb R^{n+1}$. Let us denote by $L_\lambda(U)^\perp=\{w\in\mathbb R^{n+1}\mid w^\top v=0,~\forall v\in L_\lambda(U)\}$ the orthogonal complement of $L_\lambda(U)$ (with respect to the Euclidean metric). Recall that $L_\mu(U)\subset L_\lambda^\perp(U)$ for each $\mu\in\lambda(U)-\{\lambda\}$. 

Unlike the case $U=\mathbb I_{n+1}$ considered in the previous section, the first token is no longer fixed. However, the following result shows that it converges to a fixed position provided the multiplicity of the largest eigenvalue is one.

\begin{lemma}\label{lemma:auxU}\rm
Let $\tilde\alpha:\mathbb R_0^+\to[c,\infty[$ be a continuously differentiable function for some $c\in\mathbb R^+$, and $\lambda=\max\lambda(U)$. If $\dim L_\lambda(U)=1$, then the only equilibria $\y^*\in\mathbb S^n$ of the following differential equation:
\begin{equation}\label{eq:auxU}
\dot\y=\tilde\alpha(t)(U\y-\y^\top U\y\y),\qquad\y\in\mathbb S^n,~t\in\mathbb R_0^+,
\end{equation}
are the elements $\y^*\in L_\mu(U)\cap\mathbb S^n$ for each $\mu\in\lambda(U)$. Furthermore, we have:
\begin{enumerate}
    \item $\y^*\in L_\lambda(U)\cap\mathbb S^n$ is asymptotically stable with domain of attraction: $$\mathcal D^1(\y^*)=\{\y\in\mathbb S^n\mid \y^\top\y^*>0\}.$$ 
    \item $\y^*\in L_\mu(U)\cap\mathbb S^n$, $\mu\in\lambda(U)-\{\lambda\}$, is unstable with empty domain of attraction.
\end{enumerate}
\end{lemma}

\begin{proof}
Firstly, $L_\mu(U)\cap\mathbb S^n$ is a manifold of equilibria of \cref{eq:auxU} for each $\mu\in\lambda(U)$, since:
\begin{equation*}
\tilde\alpha(t)(U\y-\y^\top U\y\y)=\mu\tilde\alpha(t)(\y-\y^\top\y\y)=0,\qquad\y\in L_\mu(U)\cap\mathbb S^n,~t\in\mathbb R_0^+.
\end{equation*}
In order to study their stability, let $\y:\mathbb R_0^+\to\mathbb S^n$ be a solution of \cref{eq:auxU} and define:
\begin{equation*}
b:\mathbb R_0^+\to[-1,1],\quad t\mapsto b(t)=v^\top\y(t),
\end{equation*}
where $v\in L_\lambda(U)\cap\mathbb S^n$. From \cref{eq:auxU} and the symmetry of $U$, the dynamics of $b$ is readily seen to be:
\begin{align*}
\dot b & =\tilde\alpha(t)v^\top (U\y-\y^\top U\y\y)\\
& =\tilde\alpha(t)(v^\top U^\top\y-\y^\top U\y v^\top\y)\\
& =\tilde\alpha(t)(\lambda v^\top\y-\y^\top U\y v^\top\y)\\
& =\tilde\alpha(t)(\lambda-\y^\top U\y)v^\top\y\\
& =\tilde\alpha(t)(\lambda-\y^\top U\y)b,
\end{align*}
By using that $\y^\top U\y\leq\lambda\y^\top\y=\lambda$ (and the equality holds if and only if $\y\in L_\lambda(U)$), the fact that $\tilde\alpha(t)>0$ for each $t\in\mathbb R_0^+$, we obtain the equilibria of the previous equation:
\begin{enumerate}
    \item $b^*=0$, which corresponds to $\y^*\in L_\lambda(U)^\perp\cap\mathbb S^n$.
    \item $b^*=-1$, which corresponds to $\y^*=-v\in L_\lambda(U)$.
    \item $b^*=1$, which corresponds to $\y^*=v\in L_\lambda(U)$.
\end{enumerate}
Moreover, $\dot b<0$ for $b\in]-1,0[$ and $\dot b>0$ for $b\in]0,1[$. Hence, $b^*=0$ is unstable with empty domain of attraction, whereas $b^*=-1$ and $b^*=1$ are asymptotically stable with domains of attraction $[-1,0[$ and $]0,1]$, respectively, which correspond to $\mathcal D^1(-v)$ and $\mathcal D^1(v)$, respectively.
\end{proof}

The previous lemma allows for establishing the asymptotic stability of two specific consensus points induced by the matrix $U$ using the same technique as in \cref{theorem:consensussytem}. Namely, the dynamics of each other token $\y_i$, $2\leq i\leq\ell$, is projected to the real line using its inner product with the corresponding asymptotically stable equilibrium of $\y_1$. By following an induction argument and treating the distance of the previous tokens to the equilibrium as an error (as it converges to zero within time), an input-to-state-stability-Lyapunov function for the projected dynamics of $\y_i$ can be found, yielding the result.

\begin{theorem}\rm\label{theorem:consensussytemU}
Suppose that \cref{ass:consensyssystemU} holds and $\dim L_\lambda(U)=1$, where $\lambda=\max\lambda(U)$. Let $v\in L_\lambda(U)\cap\mathbb S^n$. If $\lambda>0$, then $\by^*=(v,\hdots,v)$ is an asymptotically stable equilibrium of \cref{Transformer} and its domain of attraction contains the set:
\begin{align*}
\mathcal D^\ell(v)=\{(\y_1,\hdots,\y_\ell)\in(\mathbb S^n)^\ell\mid v^\top\y_i>0,~1\leq i\leq\ell\}.
\end{align*}
\end{theorem}

\begin{proof}
Firstly, the dynamics of the subsystem of \cref{Transformer} under \cref{ass:consensyssystemU} given by the first token reads: $\dot\y_1=\tilde\alpha_{11}(t)(U\y_1-\y_1^\top U\y_1\y_1)$, with $\tilde\alpha_{ij}:\mathbb R_0^+\to\mathbb R_0^+$ as in \cref{eq:tildealpha}. \cref{lemma:aux1,lemma:auxU} ensure that the result holds for that subsystem.

Now we proceed by induction: given $2\leq i\leq\ell$, the result is assumed to hold for the subsystem of \cref{Transformer} under \cref{ass:consensyssystemU} given by the first $i-1$ tokens. Hence, for each solution $(\y_1,\hdots,\y_i):\mathbb R_0^+\to(\mathbb S^n)^i$ of the subsystem of \cref{Transformer} under \cref{ass:consensyssystemU} given by the first $i$ tokens, we have that $\lim_{t\to\infty}\y_j(t)=v$ for each $1\leq j\leq i-1$ provided $(\y_1(0),\hdots,\y_{i-1}(0))\in\mathcal D^{i-1}(v)$. In order to study the behavior of the $i$-th token, we define:
\begin{align*}
e_j:\mathbb R_0^+\to\mathbb R^{n+1},\qquad & t\mapsto e_j(t)=\y_j(t)-v,\qquad 1\leq j\leq i-1,\\
b_i:\mathbb R_0^+\to[-1,1],\qquad &t\mapsto b_i(t)=v^\top\y_i(t).
\end{align*}
A straightforward check using that $U$ is symmetric and $v\in\mathbb S^n$, as well as the previous definitions, leads to:
\begin{align}\nonumber
\dot b_i & =v^\top \dot\y_i\\\nonumber
& =\sum_{j=1}^{i-1}\tilde\alpha_{ij}(t)(v^\top U(e_j+v)-\y_i^\top U(e_j+v)v^\top\y_i)+\tilde\alpha_{ii}(t)(v^\top U\y_i-\y_i^\top U\y_i v^\top\y_i)\\\nonumber
& =\sum_{j=1}^{i-1}\tilde\alpha_{ij}(t)(\lambda v^\top e_j+\lambda-(\y_i^\top U e_j+\lambda b_i)b_i)+\tilde\alpha_{ii}(t)(\lambda-\y_i^\top U\y_i)b_i\\\label{eq:dotbi}
& =\sum_{j=1}^{i-1}\tilde\alpha_{ij}(t)\lambda(1-b_i^2)+\tilde\alpha_{ii}(t)(\lambda-\y_i^\top U\y_i)b_i+g(t,e),
\end{align}
where $g$ is given by:
\begin{equation*}
g(t,e)=\sum_{j=1}^{i-1}\tilde\alpha_{ij}(t)(\lambda v^\top-b_i\y_i^\top U)e_j,\qquad e=(e_1,\hdots,e_{i-1}),
\end{equation*}
and satisfies:
\begin{align}\nonumber
|g(t,e)| & \leq\sum_{j=1}^{i-1}\tilde\alpha_{ij}(t)(\lambda|v|+|b_i|\,|\y_i|\,\|U\|)|e_j|\\\nonumber
& \leq\sum_{j=1}^{i-1}c_2(\lambda+\|U\|)|e_j|\\\label{eq:bounderror}
& =C_2\sum_{j=1}^{i-1}|e_j|,\qquad t\in\mathbb R_0^+,
\end{align}
with $c_2>0$ as in \cref{lemma:aux1} and $C_2=c_2(\lambda+\|U\|)$, where we have used that $\lambda>0$.

Given that $\lambda>0$ and $\y_i^\top U\y_i\leq\lambda\y^\top\y=\lambda$, the only equilibria of \cref{eq:dotbi} when $e=0$ are $b_i^*=1$ (which corresponds to $\y_i^*=v$) and $b_i^*=-1$ (which corresponds to $\y_i^*=-v$). In order to analyze the stability of the former, let us consider the function $V_+:[-1,1]\to\mathbb R$ defined as $V_+(b_i)=1-b_i$. We have that $V_+(1)=0$ and $V_+|_{]0,1[}>0$, as well as:
\begin{align*}
\dot V_+(t,b_i) & =V_+'(b_i)\dot b_i\\
& =-\sum_{j=1}^{i-1}\tilde\alpha_{ij}(t)\lambda(1-b_i^2)-\tilde\alpha_{ii}(t)(\lambda-\y_i^\top U\y_i)b_i-g(t,e)\\
& \leq-\tilde\alpha_{ii}(t)b_i(\lambda-\y_i^\top U\y_i)+C_2\sum_{j=1}^{i-1}|e_j|,
\end{align*}
where we have used \cref{eq:bounderror}, $\tilde\alpha_{ij}(t),\lambda>0$ for each $t\in\mathbb R_0^+$, and $0<b_i\leq 1$. For $e=0$ and $t\in\mathbb R_0^+$, $\dot V_+(t,1)=0$ and $\dot V_+(t,\cdot)|_{]0,1[}<0$. This ensures that $V_+$ is a strict ISS-Lyapunov function for the equilibrium $b_i^*=1$ of \cref{eq:dotbi}, where the input is given by $e=(e_1,\hdots,e_{i-1})$. We conclude that $b_i^*=1$ is ISS-stable with domain of attraction $]0,1]$. This corresponds to $\y_i^*=v$ with domain of attraction $\{\y_i\in\mathbb S^n\mid v^\top\y_i>0\}$. 

Lastly, if we regard the errors as function of the initial conditions i.e., $e(t)=e(t,\y_1^0,\hdots,\y_{i-1}^0)$, then the induction hypothesis ensures that $e^*=0$ is an asymptotically stable equilibrium and its domain of attraction contains $\mathcal D_1^{i-1}(v)$. As a result, $(e^*,\y_i^*)=(0,v)$ is asymptotically stable for the cascade system $(e,\y_i)$ and its domain of attraction contains $\mathcal D_1^i(v)$ (cf. Lemma~4.7 of~\cite{Kh2002}).
\end{proof}

\begin{remark}[Closest results available in the literature]
The authors were not able to find results in the literature addressing the case where $U$ is not the identity matrix although two conjectures are proposed, but not proved, in~\cite{KaPoRi2024}.
\end{remark}


\subsection{Time-varying value matrix}\label{sec:timevaryingU}

Lastly, we extend the result in the previous section to time-varying value matrices. In this case, the eigenvectors corresponding to the maximum eigenvalue of the value matrix are also time-varying. Our result states that the tokens converge to some neighborhood of this time-varying eigenvalue if they start on the hemisphere defined by the eigenvalue at the initial time. 

\begin{hypothesis}\label{ass:timevaryingU}
The model \cref{Transformer} is auto-regressive, \mbox{$W=\mathbb I_{n+1}$}, and we have:
\begin{enumerate}
    \item\label{U1} There is only one head, i.e., $h=1$, 
    \item\label{U2} $U_1(t)=U(t)$ is differentiable (as a function of $t$), bounded and symmetric, i.e., $\sup_{t\in\mathbb R_0^+}\|U(t)\|<\infty$ and $U^\top(t)=U(t)$, and
    \item\label{U3} $P_1(t)=P(t)$ is bounded, i.e., $\sup_{t\in\mathbb R_0^+}\|P(t)\|<\infty$.
\end{enumerate}
\end{hypothesis}

For each $t\in\mathbb R_0^+$, we denote the spectrum of $U(t)$ by $\lambda(U(t))=\{\lambda_1(t),\hdots,\lambda_r(t)\}$ with $\lambda_\mu(t)>\lambda_\nu(t)$ for each $1\leq\mu<\nu\leq r$. Note that $\lambda(U(t))\subset\mathbb R$ as $U(t)$ is symmetric. Given $\lambda(t)\in\lambda(U(t))$, the corresponding eigenspace is denoted by $L_{\lambda(t)}(U(t))\subset\mathbb R^{n+1}$.

Let us introduce a family of functions that will be useful for the next result. An \emph{almost class $\mathcal K\mathcal L$} function is continuous function $\Theta:\mathbb R_0^+\times\mathbb R_0^+\to\mathbb R_0^+$ such that, for each $(r_0,t_0)\in\mathbb R_0^+\times\mathbb R_0^+$, we have:
\begin{enumerate}
    \item the function $\mathbb R_0^+\ni r\mapsto\Theta(r,t_0)\in\mathbb R_0^+$ is strictly increasing, and
    \item there exists $\alpha\in\mathcal K$ such that $\lim_{t\to\infty}\Theta(r_0,t)=\alpha(r_0)$.
\end{enumerate}

\begin{theorem}\rm\label{theorem:timevaryingU}
For each $t\in\mathbb R_0^+$, let $v_1(t)\in L_{\lambda_1(t)}(U(t))\cap\mathbb S^n$ and define $\varepsilon=\sup_{t\in\mathbb R_0^+}\{|\dot v_1(t)|\}$. If \cref{ass:timevaryingU} holds and:
\begin{enumerate}
    \item\label{Ll0} $\lambda_1(t)>0$ for each $t\in\mathbb R_0^+$,
    \item\label{Ll1} $\dim L_{\lambda_1(t)}(U(t))=1$ for each $t\in\mathbb R_0^+$, and
    \item\label{Ll2} $\delta=\inf_{t\in\mathbb R_0^+}(\lambda_1(t)-\lambda_2(t))>0$, 
\end{enumerate}
then, for each $1\leq i\leq\ell$, there exist:
\begin{enumerate}
    \item a class $\mathcal{KL}$ function $\beta_i:\mathbb R_0^+\times\mathbb R_0^+\to\mathbb R_0^+$, and
    \item an almost class $\mathcal{K}\mathcal L$ function $\Theta_i:\mathbb R_0^+\times\mathbb R_0^+\to\mathbb R_0^+$,,
\end{enumerate}
such that, for each solution $\by=(\y_1,\hdots,\y_\ell):\mathbb R_0^+\to(\mathbb S^n)^\ell$ of \cref{Transformer} with $\by(0)\in\mathcal D^\ell(v_1(0))$, we have:
\begin{equation*}
|\y_i(t)-v_1(t))|^2\leq\beta_i(|\y_i(0)-v_i(0)|^2,t)+\Theta_i\left(\varepsilon,t\right),\quad 1\leq i\leq\ell,~t\in\mathbb R_0^+.
\end{equation*}
\end{theorem}

\begin{proof}
Firstly, note that \cref{U2} of \cref{ass:timevaryingU} and \cref{Ll1} in the Theorem statement ensure that $\dot v_1:\mathbb R_0^+\to\mathbb R^{n+1}$ is defined almost everywhere. For each $t\in\mathbb R_0^+$, let $\{v_1(t),\hdots,v_{n+1}(t)\}$ be an orthonormal basis of eigenvectors of $U(t)$, which exists due to its symmetry. Let us denote by $\gamma_\mu(t)\in\mathbb R$ the eigenvalue corresponding to $v_\mu(t)$ for $1\leq\mu\leq n+1$. Note that $\gamma_1(t)=\lambda_1(t)$. 

For each $1\leq i\leq\ell$, let $e_i=\y_i-v_1:\mathbb R_0^+\to\mathbb R^{n+1}$ and $b_i=v_1^\top\y_i:\mathbb R_0^+\to[-1,1]$. Henceforth, we drop the argument $t$ for simplicity. By writing $\y_i=\sum_{\mu=1}^{n+1}(v_\mu^\top\y_i)v_\mu$ we obtain $1=|\y_i|^2=\sum_{\mu=1}^{n+1}(v_\mu^\top\y_i)^2$. This, as well as the fact that $U=\sum_{\mu=1}^{n+1}\gamma_\mu\,v_\mu v_\mu^\top$, yields:
\begin{align}\nonumber
\lambda_1-\y_i^\top U\y_i & =\lambda_1-\sum_{\mu=1}^{n+1}\gamma_\mu(v_\mu^\top\y_i)^2\\\nonumber
& =\lambda_1-\lambda_1\left(1-\sum_{\mu=2}^{n+1}(v_\mu^\top\y_i)^2\right)-\sum_{\mu=2}^{n+1}\gamma_\mu(v_\mu^\top\y_i)^2\\\nonumber
& =\sum_{\mu=2}^{n+1}(\lambda_1-\gamma_\mu)(v_\mu^\top\y_i)^2.
\end{align}
From this and \cref{Transformer} under the hypothesis in \cref{ass:timevaryingU}, the dynamics of $b_1$ reads:
\begin{align}\nonumber
\dot b_1 & =v_1^\top \dot\y_1 +\dot v_1^\top \y_1 \\\nonumber
& =\tilde\alpha_{11} \left(v_1^\top U \y_1 -\y_1^\top U \y_1 v_1^\top \y_1 \right)+\dot v_1^\top \y_1 \\\nonumber
& =\tilde\alpha_{11} \left(\lambda_1 -\y_1^\top U \y_1 \right)b_1+\dot v_1^\top \y_1 ,\\\label{eq:dotb1U}
& =\tilde\alpha_{11} b_1\sum_{\mu=2}^{n+1}(\lambda_1 -\gamma_\mu )(v_\mu^\top\y_1)^2 +\dot v_1^\top \y_1 ,
\end{align}
for each $(t,b_1)\in\mathbb R_0^+\times[-1,1]$. Similarly, the dynamics of $b_i$, $2\leq i\leq\ell$, reads:
\begin{align}\nonumber
\dot b_i & =v_1^\top\dot\y_i+\dot v_1^\top\y_i\\\nonumber
& =\sum_{j=1}^{i-1}\tilde\alpha_{ij}(v_1^\top U(e_j+v_1)-\y_i^\top U(e_j+v_1)v_1^\top\y_i)+\tilde\alpha_{ii}(v_1^\top U\y_i-\y_i^\top U\y_i v_1^\top\y_i)+\dot v_1^\top\y_i\\\nonumber
& =\sum_{j=1}^{i-1}\tilde\alpha_{ij}(\lambda_1 v_1^\top e_j+\lambda_1-(\y_i^\top U e_j+\lambda_1 b_i)b_i)+\tilde\alpha_{ii}(\lambda_1-\y_i^\top U \y_i)b_i+\dot v_1^\top\y_i\\\nonumber
& =\sum_{j=1}^{i-1}\tilde\alpha_{ij}\lambda_1(1-b_i^2)+\tilde\alpha_{ii}(\lambda_1-\y_i^\top U\y_i)b_i+g(e)+\dot v_1^\top\y_i,\\\label{eq:dotbiU}
& =\sum_{j=1}^{i-1}\tilde\alpha_{ij}\lambda_1(1-b_i^2)+\tilde\alpha_{ii}b_i\sum_{\mu=2}^{n+1}(\lambda_1-\gamma_\mu)(v_\mu^\top\y_i)^2+g(e)+\dot v_1^\top\y_i,
\end{align}
for each $(t,b_i)\in\mathbb R_0^+\times[-1,1]$, where $e=(e_1,\hdots,e_{i-1})$ and $g(e)=\sum_{j=1}^{i-1}\tilde\alpha_{ij}(\lambda_1 v_1^\top-b_i\y_i^\top U)e_j$. Note that:
\begin{align}\nonumber
|g(e)| & \leq\sum_{j=1}^{i-1}\tilde\alpha_{ij}(\lambda_1|v_1|+|b_i|\,|\y_i|\,\|U\|)|e_j|\\\nonumber
& \leq\sum_{j=1}^{i-1}c_2(\sup_{t\in\mathbb R_0^+}\lambda_1(t)+\sup_{t\in\mathbb R_0^+}\|U(t)\|)|e_j|\\\label{boundg(e)}
& =C\sum_{j=1}^{i-1}|e_j|,
\end{align}
with $c_2>0$ as in \cref{lemma:aux1} and $C=c_2(\sup_{t\in\mathbb R_0^+}\lambda_1(t)+\sup_{t\in\mathbb R_0^+}\|U(t)\|)\in\mathbb R^+$, where we used \cref{U2} and \cref{Ll0} to ensure that $0<\sup_{t\in\mathbb R_0^+}\lambda_1(t)<\infty$. 

Let us proceed by complete induction. 
\begin{enumerate}
    \item\textbf{Base case.} For $i=1$, \cref{Ll1} and the orthogonality of the basis of eigenvectors chosen ensure that the only equilibria of \cref{eq:dotb1U} are $b_1^*=-1$ (which corresponds to $\y_1^*=-v_1$) and $b_1^*=1$ (which corresponds to $\y_1^*=v_1)$, when $\dot v_1=0$. In order to analyze the stability of the latter, let us consider the function $V_+:[-1,1]\to\mathbb R$ defined as $V_+(b_1)=1-b_1$. Clearly, $V_+(1)=0$ and $V_+|_{]0,1[}>0$. Moreover, from \cref{Ll1,Ll2} and \cref{eq:dotb1U}, we obtain:
    \begin{align*}
    \dot V_+(t,b_1) & =V_+'(b_1)\,\dot b_1\leq-\delta c_1(1-b_1^2)b_1+\varepsilon,\qquad(t,b_1)\in\mathbb R_0^+\times [0,1],
    \end{align*}
    with $c_1>0$ as in \cref{lemma:aux1}. For $\varepsilon=0$ and $t\in\mathbb R_0^+$, $\dot V_+(t,1)=0$ and $\dot V_+(t,\cdot)|_{]0,1[}<0$. This ensures that $V_+$ is a strict ISS-Lyapunov function for the equilibrium $b_1^*=1$ of \cref{eq:dotb1U}, where the input is given by $\dot v_1^\top\y_1$. We conclude that $b_1^*=1$ is ISS-stable with domain of attraction $]0,1[$. Hence, there exist $\tilde\alpha_1\in\mathcal K$ and $\tilde\beta_1\in\mathcal{KL}$ such that, for each solution $b_1:\mathbb R_0^+\to[-1,1]$ of \cref{eq:dotb1U} with $b_1(0)\in]0,1]$ (which corresponds to $\y_1(0)\in\mathcal D(v_1(0))$), we have $|b_1(t)-1|\leq\tilde\beta_1(|b_1(0)-1|,t)+\tilde\alpha_1(\varepsilon)$ for each $t\in\mathbb R_0^+$. By noting that $|\y_1-v_1|=|e_1|=\sqrt{\langle e_1,e_1\rangle}=\sqrt{2|b_1-1|}$, the previous condition may be rewritten as:
    \begin{align*}
    |\y_1(t)-v_1(t)|^2 & =2|b_1(t)-1|\\
    & \leq2\tilde\beta_1(|b_1(0)-1|,t)+2\tilde\alpha_1(\varepsilon)\\
    & =2\tilde\beta_1\left(\frac{|y_1(0)-v_1(0)|^2}{2},t\right)+2\tilde\alpha_1(\varepsilon),
    \end{align*}
    for each $t\in\mathbb R_0^+$. By picking $\beta_1(r,t)=2\tilde\beta_1(r/2,t)$ and $\Theta_1(r,t)=2\tilde\alpha_1(r)$ for each $(r,t)\in\mathbb R_0^+\times\mathbb R_0^+$, we conclude.  
    
    \item\textbf{Inductive step.} Now, for $2\leq i\leq\ell$ let us assume that the result holds for the subsystem of \cref{Transformer} under the assumptions in the statement given by the first $i-1$ tokens. Namely, there exist a class $\mathcal{KL}$ function $\beta_j:\mathbb R_0^+\times\mathbb R_0^+\to\mathbb R_0^+$ and an almost class $\mathcal{KL}$ function $\Theta_j:\mathbb R_0^+\times\mathbb R_0^+\to\mathbb R_0^+$, $1\leq j\leq i-1$, such that, for each solution $(\y_1,\hdots,\y_j):\mathbb R_0^+\to(\mathbb S^n)^{i-1}$ of such subsystem with $(\y_1(0),\hdots,\y_j(0))\in\mathcal D^j(v_1(0))$, we have:
    \begin{equation}\label{eq:ISSyjU}
    |e_j(t)|^2\leq\beta_j(|e_j(0)|^2,t)+\Theta_j(\varepsilon,t),\qquad t\in\mathbb R_0^+,~1\leq j\leq i-1.
    \end{equation}
    Thanks to the orthogonality of the basis of eigenvectors, the only equilibria of \cref{eq:dotbiU} are $b_i^*=-1$ (which corresponds to $\y_i^*=-v_1$) and $b_i^*=1$ (which corresponds to $\y_i^*=v_1$) when $e=(e_1,\hdots,e_{i-1})=0$ and $\dot v_1=0$. To analyze the stability of the latter, let us consider the function $V_+:[-1,1]\to\mathbb R$ as defined above, i.e., $V_+(b_i)=1-b_i$, which satisfies $V_+(1)=0$ and $V_+|_{]0,1[}>0$. From \cref{Ll0,Ll1,Ll2}, \cref{eq:dotbiU} and \cref{boundg(e)}, we obtain:
    \begin{align}\label{eq:bounddotbi}
    \dot V_+(t,b_i) & =V_+'(b_i)\,\dot b_i\leq-\delta c_1(1-b_i^2)b_i+C\sum_{j=1}^{i-1}|e_j|+\varepsilon,
    \end{align}
    for each $(t,b_i)\in\mathbb R_0^+\times[0,1]$. For $e=(e_1,\hdots,e_{i-1})=0$ and $\varepsilon=0$, $\dot V_+(t,1)=0$ and $\dot V_+(t,\cdot)|_{]0,1[}<0$. Thus, $V_+$ is a strict ISS-Lyapunov function for the equilibrium $b_i^*=1$ of \cref{eq:dotbiU}, where the input is given by $\tilde e=C\sum_{j=1}^{i-1}|e_j|+\varepsilon$. Given that $b_i^*=1$ is ISS-stable with domain of attraction $]0,1[$, there exist $\tilde\alpha_i\in\mathcal K_\infty$ and $\tilde\beta_i\in\mathcal{KL}$ such that, for each solution $b_i:\mathbb R_0^+\to[-1,1]$ of \cref{eq:dotbiU} with $b_i(0)\in]0,1]$ (which corresponds to $\y_i(0)\in\mathcal D(v_1(0))$) we have: 
    \begin{align*}
    |b_i(t)-1| & \leq\tilde\beta_i(|b_i(0)-1|,t)+\tilde\alpha_i\left(C\sum_{j=1}^{i-1}|e_j|+\varepsilon\right),\qquad t\in\mathbb R_0^+.
    \end{align*}
    From this, the fact that $|\y_j-v_1|=|e_j|=\sqrt{2|b_j-1|}$ for $1\leq j\leq i$ and \cref{eq:ISSyjU}, we obtain:
    \begin{align*}
    & |e_i(t)|^2 =2|b_i(t)-1|\leq2\tilde\beta_i(|b_i(0)-1|,t)+2\tilde\alpha_i\left(C\sum_{j=1}^{i-1}|e_j|+\varepsilon\right)\\
    & \leq2\tilde\beta_i\left(\frac{|e_i(0)|^2}{2},t\right)+2\tilde\alpha_i\left(C\sum_{j=1}^{i-1}\sqrt{\beta_j(2,t)+\Theta_j(\varepsilon,t)}+\varepsilon\right),
    \end{align*}
    for each $t\in\mathbb R_0^+$, where we used that $|e_j(0)|\leq\sqrt{2}$ as $\y_j(0)\in\mathcal D(v_1(0))$, $1\leq j\leq i-1$. We conclude by choosing $\beta_i(r,t)=2\tilde\beta_i(r/2,t)$ and:
    \begin{equation*}
    \Theta_i(r,t)=2\tilde\alpha_i\left(C\sum_{j=1}^{i-1}\sqrt{\beta_j(2,t)+\Theta_j(r,t)}+r\right).
    \end{equation*}
\end{enumerate}
\end{proof}

Under the conditions of \cref{theorem:timevaryingU} 
, there exist $\alpha_i\in\mathcal K$ such that:
\begin{equation*}
\lim_{t\to\infty}|\y_i(t)-v_1(t)|^2\leq\alpha_i(\varepsilon),\qquad 1\leq i\leq\ell,
\end{equation*}
provided $\by(0)=(\y_1(0),\hdots,\y_\ell(0))\in\mathcal D^\ell(v_1(0))$. Therefore, all tokens converge to some ball around the time-varying consensus $\by^*(t)=(v_1(t),\hdots,v_1(t))$ as $t\to\infty$. The radius of the ball depends on $\varepsilon=\sup_{t\in\mathbb R_0^+}\{|\dot v_1(t)|\}$. Moreover, the basin of attraction is the open hemisphere generated by $v_1(0)$. For the limit case where $\varepsilon=0$, i.e., when $v_1(t)$ does not depend on time, we recover exact convergence to the consensus, as in \cref{theorem:consensussytemU}.

\begin{remark}[Closest results available in the literature]
The authors are not aware of any previous results addressing the case where the value matrix is time-varying.
\end{remark}

\section{Simulations and empirical validation}
\label{Sec:Simulations}



In this section we illustrate the theoretical results and show that their conclusions appear to hold even when our assumptions are violated.
We start by simulating the continuous transformer model and illustrating our theoretical results. In addition to simulations, we provide empirical evidence using the GPT-2 XL and the GPT-Neo 2.7B to show how token consensus seems to occur even if the assumptions in our theoretical results are not satisfied.

\subsection{Numerical simulations}
\subsubsection{Illustration of Theorem \ref{theorem:hemisphere}}
\label{SSec:Illustration1}
We simulate the motion of 10 tokens, each of them randomly placed on the sphere $\mathbb S^2\subset\mathbb R^3$, according to the dynamics~\cref{Transformer} with $h=2$. All matrices, except for $P_1(t)$ and $P_2(t)$, were randomly chosen, and each element was drawn from a uniform distribution in the interval $[-0.5, 0.5]$. The matrices $P_1(t)$ and $P_2(t)$ were computed as $P_1(t)=D_1(t)P_1'$, $P_2(t)=D_2(t)P_2'$ with $P_1'$ and $P_2'$ randomly generated:
\begin{align*}
& P_1' = 
\begin{pmatrix}
0.08 & -0.19 & 0.20 \\
-0.23 & 0.31 & -0.23 \\
0.18 & -0.17 & -0.16
\end{pmatrix},\qquad P_2' = 
\begin{pmatrix}
-0.31 & 0.03 & 0.11 \\
0.06 & -0.06 & 0.13 \\
0.14 & 0.11 & 0.10
\end{pmatrix}.
\end{align*}
The matrices $D_1(t)$ and $D_2(t)$ were given by:
\begin{align*}
D_1(t) & =2\operatorname{diag}\big(\cos(10 \pi t),~\sin(10 \pi t),~\cos(6 \pi t)\big),\\
D_2(t) & =2\operatorname{diag}\big(\cos(6 \pi t),~\sin(6 \pi t),~\cos(4 \pi t)\big),    
\end{align*}
where \mbox{$\operatorname{diag}:\mathbb R^3\to\mathcal{M}_{3\times3}(\mathbb R)$} denotes the function that maps a vector to the diagonal matrix with its components on the diagonal.

\begin{figure}[ht]
    \centering{
        \includegraphics[width=0.45\textwidth, keepaspectratio]{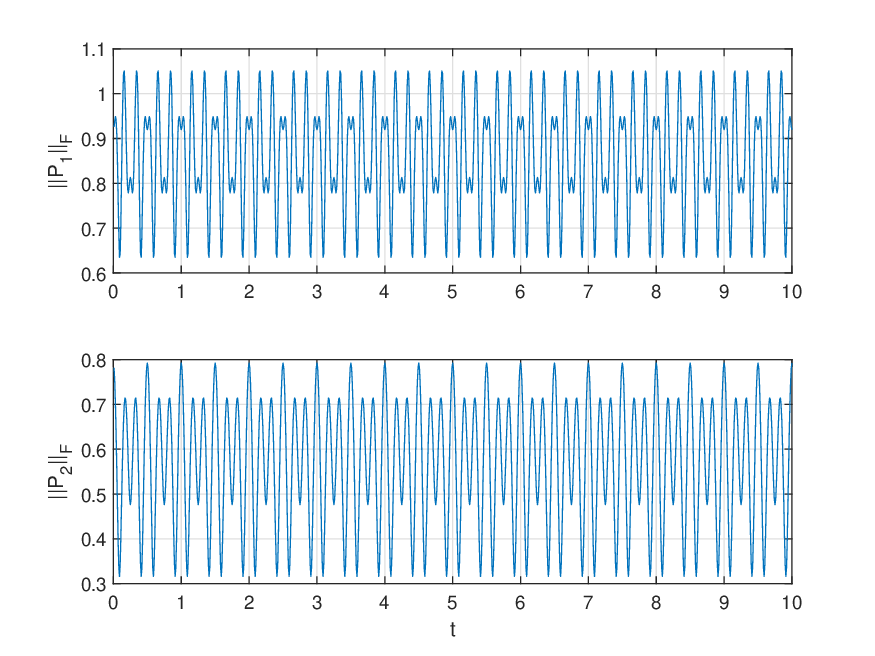}}
    \caption{Frobenius norm of the matrices $P_1(t)$ and $P_2(t)$.}
    \label{norm}
\end{figure}

 \begin{figure}[ht]
    \centering
    {\includegraphics[width=0.45\textwidth, keepaspectratio]{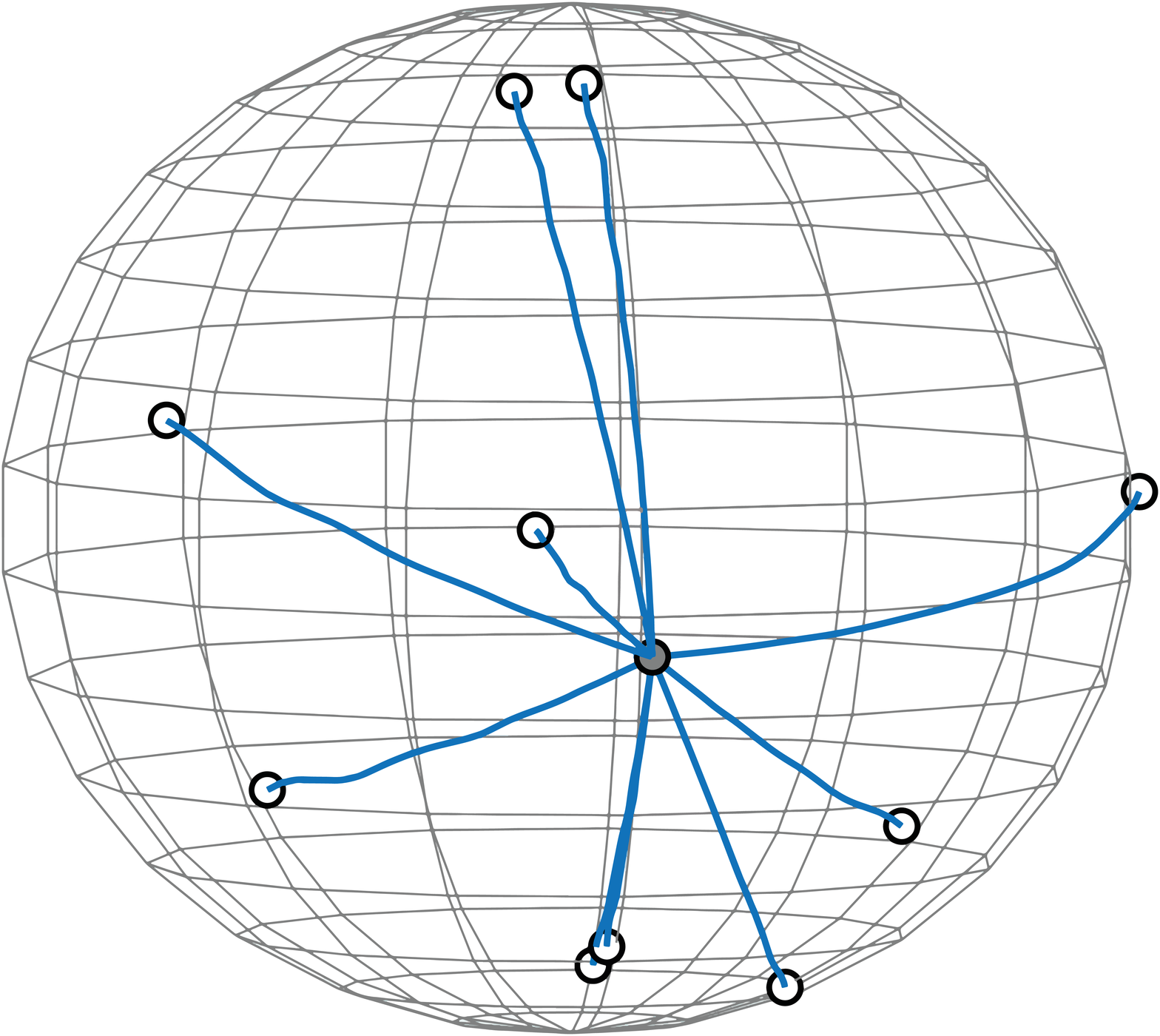}}
    \caption{Convergence  to a consensus equilibrium on the sphere $\mathbb S^2$. All the tokens start and remain in an hemisphere.\label{4:2.1}}
    \end{figure}

To better appreciate the time-varying nature of the matrices $P_1$ and $P_2$, in \cref{norm} we shown their Frobenius norm.

In \cref{4:2.1} we show the motion of the tokens in blue with their initial position represented by a white circle and final position by a gray circle. We can appreciate that all the tokens start and remain in an hemisphere and that they converge to a consensus equilibrium.

\begin{figure}[t]
    \centering      {\includegraphics[width=0.45\textwidth,keepaspectratio]{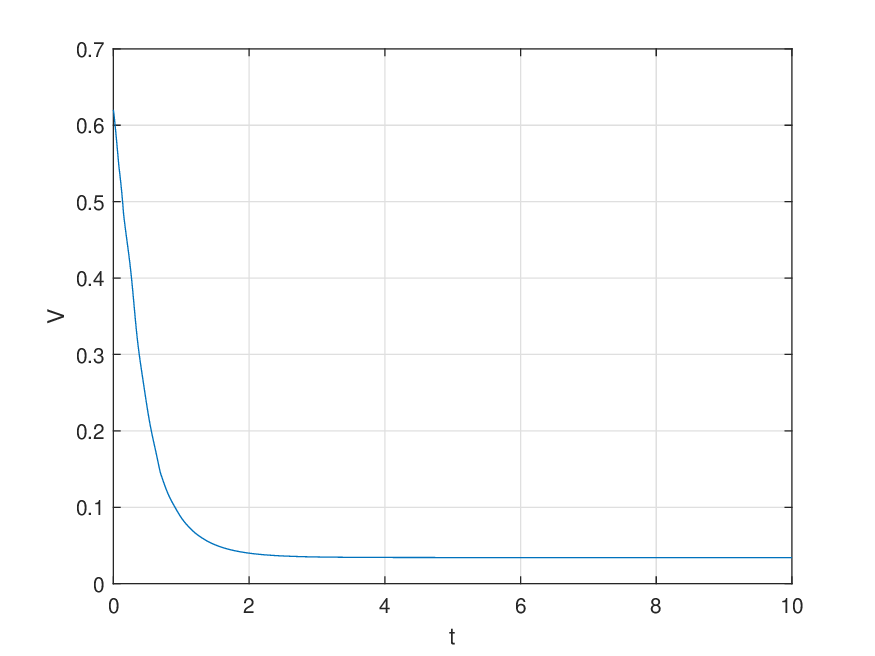} 
   \caption{Evolution of the Lyapunov function~\cref{LyapunovFunction} used in the proof of \cref{theorem:hemisphere}.}\label{4:2.2}}  
\end{figure}

The proof of \cref{theorem:hemisphere} is based on the Lyapunov function~\cref{LyapunovFunction}, whose time-evolution is displayed in \cref{4:2.2} for the case where $v=(1,0,0)$.

\subsubsection{Illustration of \texorpdfstring{\cref{theorem:consensussytem}}{Theorem 5.3}}
We now consider the auto-regressive model with $50$ tokens on $\mathbb S^{499}\subset\mathbb R^{500}$. The number and dimension of the tokens were chosen to make them comparable to the GPT-2 model. We use two heads ($h=2$) with the matrices $P_1=D_1(t)P_1'$ and $P_2=D_2(t)P_2'$ obtained by randomly generating $P_1'$ and $P_2'$, and taking $D_1(t)$ and $D_2(t)$ to be diagonal with entries $(D_\eta)_{jj}=|2\sin(w t + \phi)|$ for $\eta=1,2$, $j=1,\hdots,500$, $w$ drawn from the uniform distribution on $]0,1[$ and $\phi$ drawn from the uniform distribution on $]0, 2\pi[$. To measure the error between tokens we use the cosine similarity, $E : (\mathcal{E}_W^n)^\ell \to \mathbb{R}^+$, defined as: \begin{equation}
\label{E}
    E =  1-\frac{1}{\ell}\sum_{i=1}^\ell \frac{\y_1^\top\y_i}{|\y_1|\,|\y_i|},
\end{equation}
which becomes zero when all the tokens belong to the consensus set. 

\begin{figure}[ht]
    \centering{
        \includegraphics[width=0.45\textwidth, keepaspectratio]{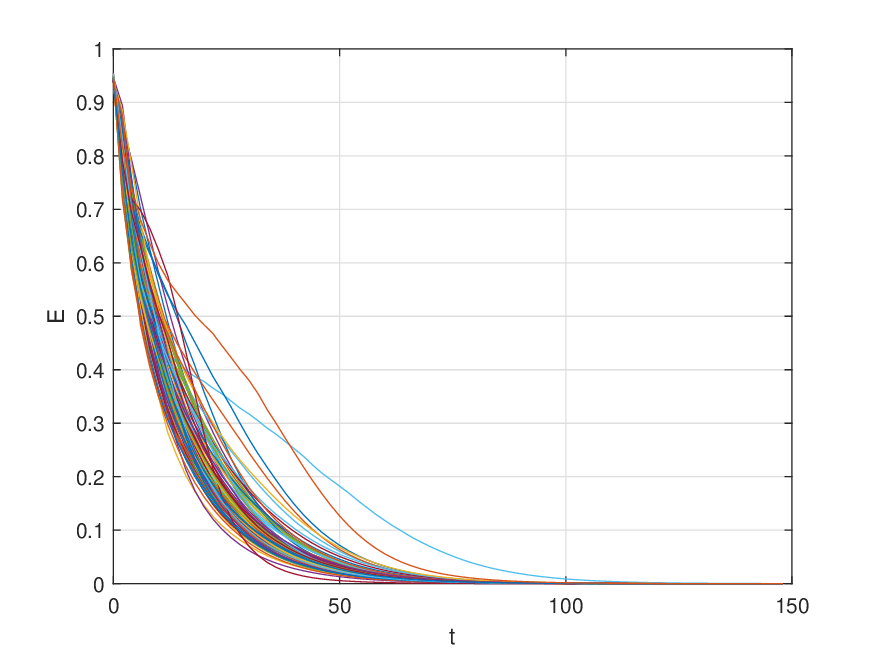}}
    \caption{Illustration of \cref{theorem:consensussytem}; evolution of the function $E$ defined in~\cref{E} along $100$ solutions of~\cref{Transformer} with random initial conditions drawn from an element-wise uniform distribution on $]-0.5,0.5[$ and then projected to the sphere.}
    \label{4:4}
\end{figure}

In \cref{4:4} we display the evolution of the function $E$ along $100$ trajectories of~\cref{Transformer} for random initial conditions drawn from an element-wise uniform distribution on $]-0.5,0.5[$, and then projected to the sphere. 
We can appreciate in \cref{4:4} how the function $E$ converges to zero along all the trajectories.

\subsubsection{Illustration of \texorpdfstring{\cref{theorem:consensussytemU}}{Theorem 5.8}}

In the final case we use the causal model with $10$ tokens on $\mathbb S^2$ with randomly assigned initial positions. As for the previous cases, we choose $P(t)=D(t)P'$, with randomly generated $P'$ and $U$ given by:
\begin{align*}
 P' = \begin{pmatrix}
0.36 & 0.42 & 0.13 \\
0.10 & -0.07 & -0.20 \\
0.15 & -0.21 & 0.12
\end{pmatrix},\qquad
 U = \begin{pmatrix}
-0.26 & 0.50 & 0.56 \\
0.50 & -0.72 & -0.50 \\
0.56 & -0.50 & -0.02
\end{pmatrix},
\end{align*}
and $D(t)=2\operatorname{diag}\big(\cos(10 \pi t),~\sin(10 \pi t),~\cos(6 \pi t)\big)$.

\begin{figure}[ht]
    \centering
    {\includegraphics[width=0.45\textwidth, keepaspectratio]{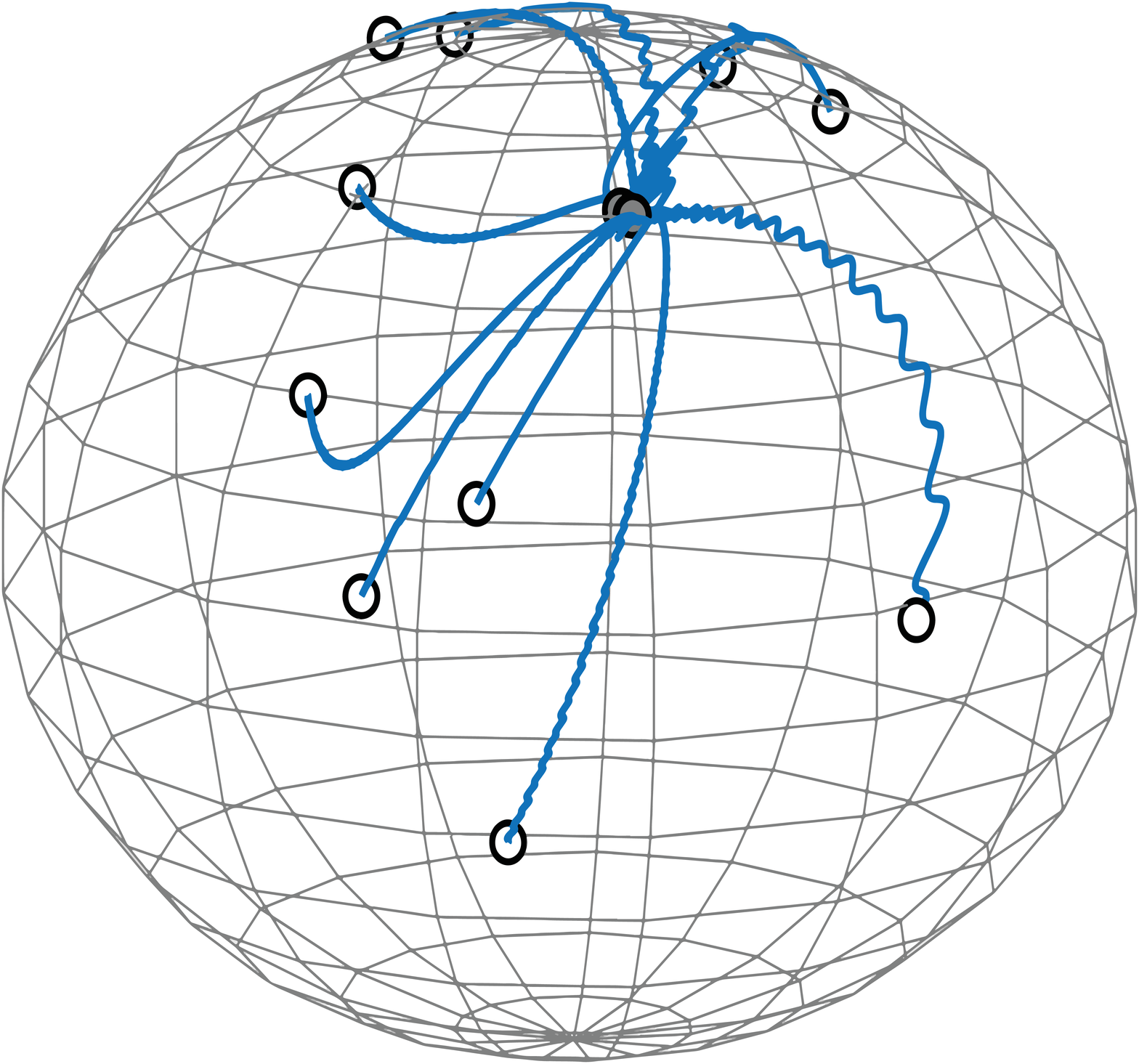}}\caption{Convergence  to a consensus equilibrium on the sphere $\mathbb S^2$. All the tokens start and remain in the hemisphere defined by $v$.\label{4:3.1}}
    \end{figure}

\begin{figure}[ht]\centering{        \includegraphics[width=0.45\textwidth,keepaspectratio]{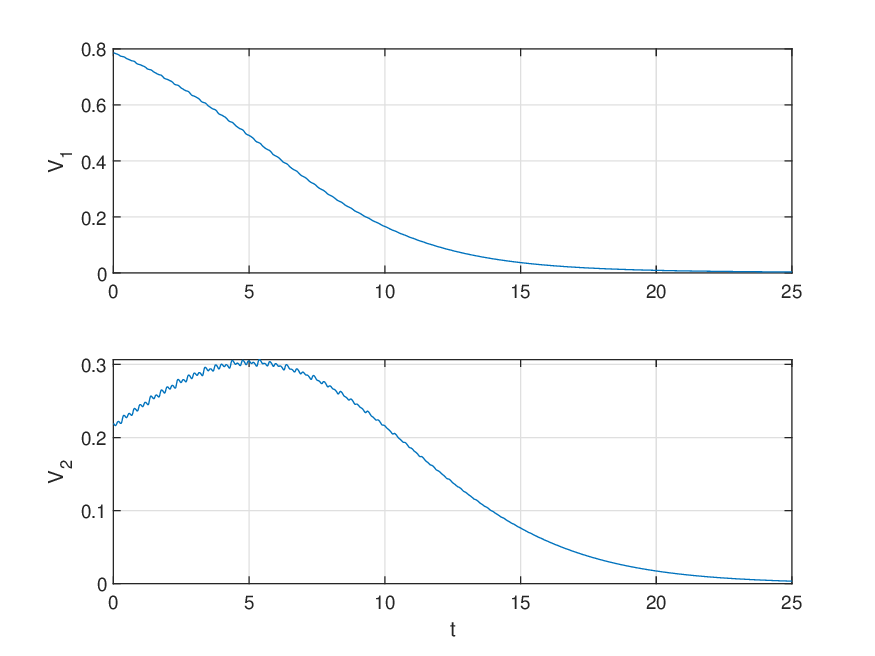}} \caption{Evolution of the functions $V_1$ and $V_2$ used in the proof of \cref{theorem:consensussytemU}.\label{4:3.2}}
\end{figure}

In \cref{4:3.1} we can observe convergence of the tokens to a consensus equilibrium point whereas in \cref{4:3.2} we have the time evolution of $V_1=1-\y_1^\top v$ and $V_2=1-\y_2^\top v$ where $v\in\mathbb{R}^3$ is the eigenvector of $U$ corresponding to its largest eigenvalue. Note that $V_2$ is not a Lyapunov function, and therefore it may increase, although the proof of \cref{theorem:consensussytemU}, establishes that it will eventually converge to zero. 

\subsection{GPT-2 and GPT-Neo Experiments} In this section we report on experiments conducted on the GPT-2 XL model and the GPT-Neo 2.7B model suggesting that our theoretical findings hold under more general assumptions. Since our results are asymptotic, we need to increase the depth of both models. We do so by running the same set of tokens through the model multiple times. In other words, we extract the tokens at the end of the model, after the final normalization, and feed them to the model for another pass thereby simulating a model of increased length, the code used for our experiments is available at~\cite{cyphylab2025gptconsensus}. 

For all our experiments, we used the same set of 100 random prompts, each generated by uniformly sampling 200 tokens from the GPT-2 tokenizer's vocabulary. In each experiment, we plot the average of $E$ across all the prompts. As an example, the first 10 tokens of the first sampled prompt are:
\begin{center}
\begin{small}
\texttt{divest anxYou coasts Oz\\Vi Happy appreciate tcp} 
\end{small}.
\end{center}

In the first experiments, we removed the feedfoward layers of each model, to make them closer to the structure we assume in our theoretical work. The experiments were then repeated  without removing the feedforward layers, showing that in both cases convergence to consensus occurs.

\begin{figure}[ht]
    \centering{
            \includegraphics[width=0.45\textwidth, keepaspectratio]{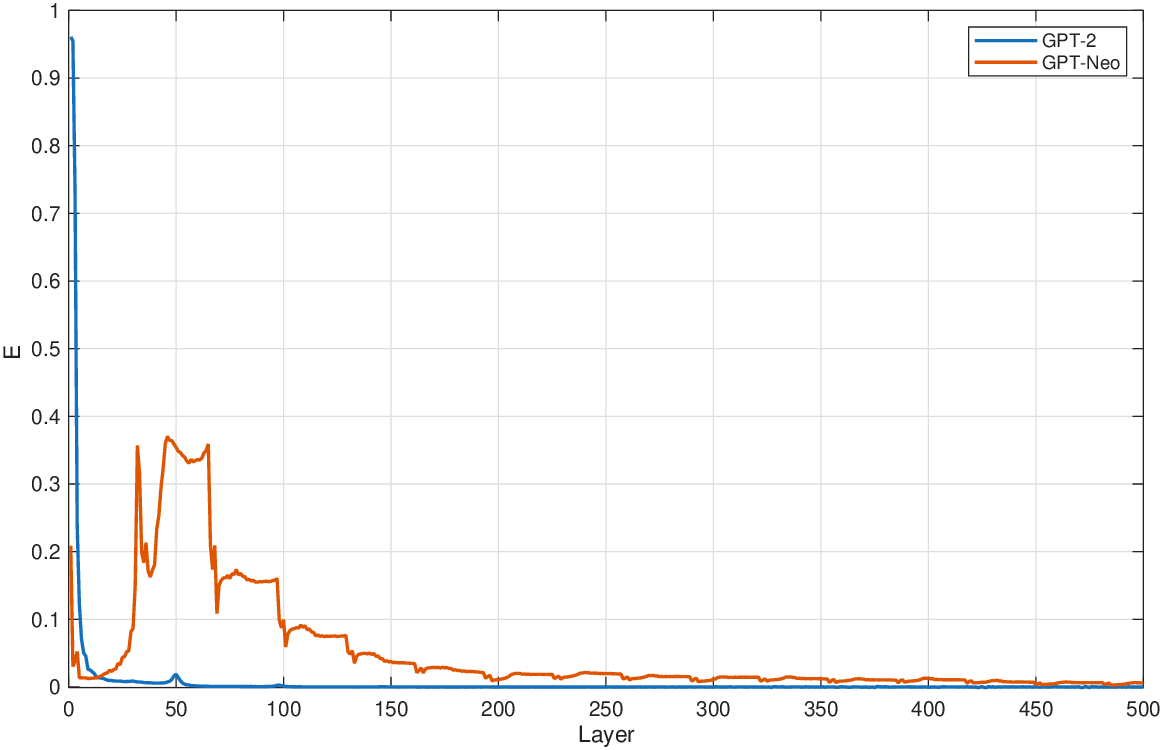}}
    \caption{Comparison between the GPT-2 XL and GPT-Neo 2.7B architectures with feedforward layers removed; evaluation of the average of~\cref{E} across all the random prompts.}
    \label{4:5}
\end{figure}

\begin{figure}[ht]
    \centering{
        \includegraphics[width=0.45\textwidth, keepaspectratio]{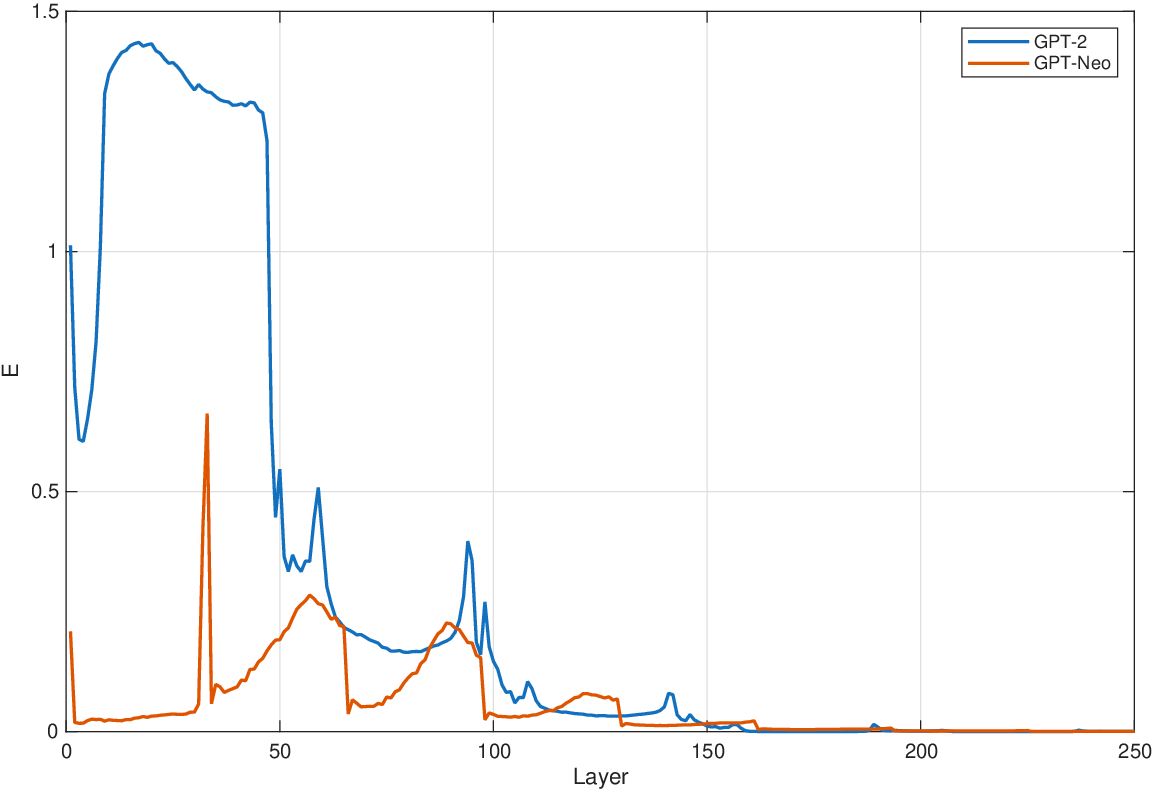}}
    \caption{Comparison between the GPT-2 XL and GPT-Neo 2.7B architectures with the full model; evaluation of the average of~\cref{E} across all the random prompts.}
    \label{4:6}
\end{figure}

The experiments were conducted on the standard configuration of the GPT-2 XL model and the GPT-Neo 2.7B model, using the pre-trained weights provided by the Hugging Face library~\cite{wolf2020transformers}. The multiple passes through the model results in matrices $P$ and $U$ that are time-varying but periodic with period corresponding to the depth each model: 48 layers for the GPT-2 XL and 32 for the GPT-Neo 2.7B. To measure how far the tokens are from each other we used the function \cref{E} whose evaluation after each layer is depicted in \cref{4:5,4:6}.

We can observe that in both models the average of the function $E$ over all the prompts converge asymptotically to 0, thus implying the tokens converge to a consensus equilibrium. We recall that our theoretical results predict this observation \textbf{only} when feedforward layers are absent, i.e., the case depicted in \cref{4:5}. However, as can be seen in \cref{4:6}, even when the feedfoward layers are present, convergence still occurs. The rate of convergence appears to be dependent on the weights of the feedfoward layers as their presence increases the convergence rate in the GPT-2 model, but decreases it in the GPT-Neo model. In both cases, these findings suggest that feedfoward layers may not be sufficient to preclude consensus.

\begin{figure}[!b]
    \centering{
        \includegraphics[width=0.45\textwidth, keepaspectratio]{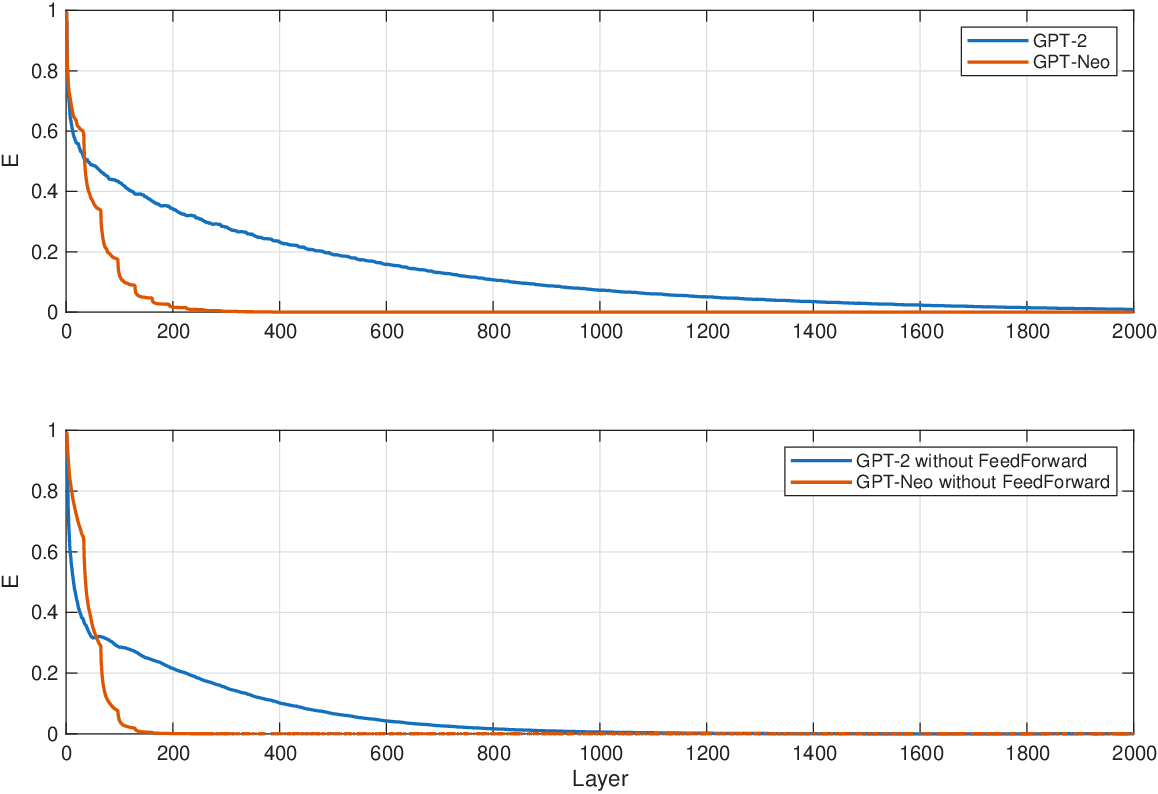}}
    \caption{Comparison between the GPT-2 XL and GPT-Neo 2.7B architectures with fixed and randomly chosen weight matrices. Each model was evaluated with and without feedfoward layers using the average of~\cref{E} across all the random prompts.}
    \label{4:7}
\end{figure}

\begin{figure}[!b]
    \centering{
        \includegraphics[width=0.45\textwidth, keepaspectratio]{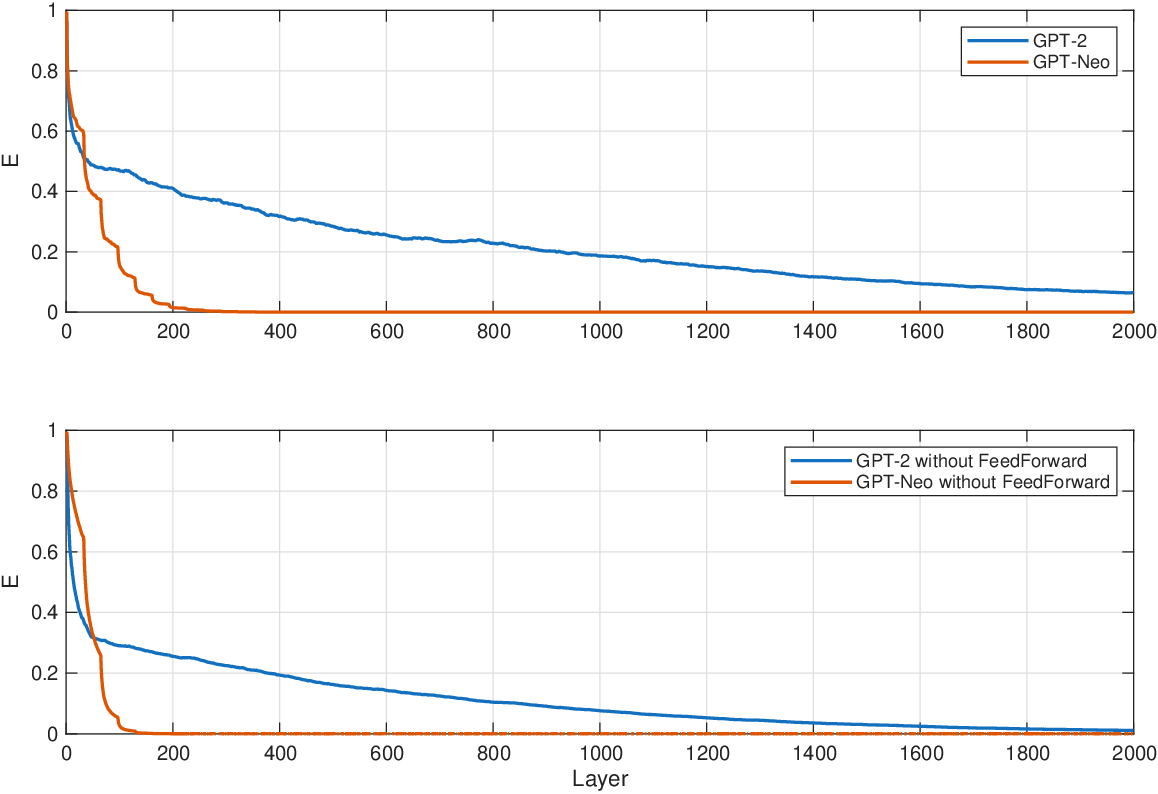}}
    \caption{Comparison between the GPT-2 XL and GPT-Neo 2.7B architectures with random weight matrices chosen before each model pass. Each model was evaluated with and without feedfoward layers using the average of~\cref{E} across all the random prompts.}
    \label{4:8}
\end{figure}

Although the previous experimental results suggest that consensus occurs even in the presence of feedfoward layers, it does not address the question of consensus being a structural property of the the transformer architecture or of the choice of weight matrices. To address this question we repeated the experiments by using random matrices in GPT-2 and GPT-Neo. Since this results in time-varying matrices, we further repeated the experiments by randomly selecting new weight matrices before each model pass. Moreover, we conducted these experiments with the full model and also by removing  the feedforward layers (including the associated normalization function and skip connection) to better understand the impact of these on token consensus. The results are reported in \cref{4:7,4:8}, where we can see that convergence towards consensus still occurs across all experiments. Furthermore, \cref{4:7,4:8} suggest that the feedfoward layer may decrease the convergence rate.

Our experiments suggest that the convergence phenomenon is a product of the structure of the transformers and not of the choice of weights. We observe convergence with trained, random periodic, and random aperiodic matrices $P$ and $U$. In terms of rate of convergence, the choice of weight matrices appears to have an impact with faster convergence being observed when pretrained matrices were used.

\bibliographystyle{ieeetr}
\bibliography{LLM.bib}

\end{document}